%% file: paper.tex
\documentclass[10pt,journal,compsoc]{IEEEtran}
\usepackage{times} % Ensures font consistency by using Times Roman
\usepackage{graphicx}
\usepackage{amsmath}
\usepackage{multirow}
\usepackage{amssymb}
\usepackage{booktabs}
\usepackage{color}
\usepackage[dvipsnames]{xcolor}
\usepackage{makecell}
\usepackage{amsfonts}
\usepackage{makecell}
\usepackage{gensymb}

\usepackage{bbding}
\usepackage{pifont}
\usepackage{wasysym}

\usepackage[pagebackref,breaklinks,colorlinks]{hyperref}
\usepackage{enumitem}
\usepackage{enumerate}
\usepackage{multirow}
\usepackage{booktabs}
\usepackage{xspace}

\makeatletter
\DeclareRobustCommand\onedot{\futurelet\@let@token\@onedot}
\def\@onedot{\ifx\@let@token.\else.\null\fi\xspace}

\def\ie{\emph{i.e}\onedot}

\makeatother

\newif\ifshowcomments
\showcommentstrue
\ifshowcomments

\newcommand{\TODO}[1]{{\color{red}{[TODO: #1]}}}

\newcommand{\revised}[1]{{\color[rgb]{0.2,0.7,0.2}{#1}}}

\else
\newcommand{\TODO}[1]{}
\newcommand{\revised}[1]{}
\fi

\ifCLASSOPTIONcompsoc
  \usepackage[nocompress]{cite}
\else
  \usepackage{cite}
\fi
\ifCLASSINFOpdf
\else
\fi
\begin{document}
\title{Demystify Transformers \& Convolutions in Modern Image Deep Networks}

\author{Xiaowei Hu, Min Shi, Weiyun Wang, Sitong Wu, Linjie Xing, Wenhai Wang, Xizhou Zhu, \\ Lewei Lu, Jie Zhou, Xiaogang Wang, Yu Qiao, and Jifeng Dai
\IEEEcompsocitemizethanks{
\IEEEcompsocthanksitem Xiaowei Hu is with South China University of Technology, Guangzhou, China.
\IEEEcompsocthanksitem Min Shi, Weiyun Wang, Sitong Wu, Linjie Xing, Wenhai Wang, and Yu Qiao are with the Shanghai Artificial Intelligence Laboratory, Shanghai, China.
\IEEEcompsocthanksitem Min Shi is also with Huazhong University of Science and Technology, Wuhan, China, Weiyun Wang is also with Fudan University, Shanghai, China, Sitong Wu is also with The Chinese University of Hong Kong, Shatin, Hong Kong.
\IEEEcompsocthanksitem Xizhou Zhu, Jie Zhou, and Jifeng Dai are with Tsinghua University, Beijing, China.
\IEEEcompsocthanksitem Lewei Lu is with SenseTime Research, Beijing, China.
\IEEEcompsocthanksitem Xiaogang Wang is with The Chinese University of Hong Kong, Shatin, Hong Kong.
\IEEEcompsocthanksitem Joint first authors: Xiaowei Hu, Min Shi, Weiyun Wang, and Sitong Wu.
\IEEEcompsocthanksitem Corresponding author: Jifeng Dai (daijifeng@tsinghua.edu.cn).
}
}

\markboth{IEEE Transactions on Pattern Analysis and Machine Intelligence}%
{Shell \MakeLowercase{\textit{et al.}}: Bare Demo of IEEEtran.cls for Computer Society Journals}
\IEEEtitleabstractindextext{%

\input{Section0-abstract}

\begin{IEEEkeywords}
Spatial token mixer, transformer, CNN, and image deep network.
\end{IEEEkeywords}
}

\maketitle

\IEEEdisplaynontitleabstractindextext

% \fi
\IEEEpeerreviewmaketitle

%%%%%%%%% BODY TEXT

\input{Section1-introduction}

\input{Section2}
\input{Section3-evolution.tex}

\input{Section4}
\input{Section5}

\input{Section6}
\input{Section7}

\if 0
\appendices
\section{Proof of the First Zonklar Equation}
Appendix one text goes here.

\section{}
Appendix two text goes here.
\fi


\ifCLASSOPTIONcaptionsoff
  \newpage
\fi

{\small
	\bibliographystyle{IEEEtran}
	\bibliography{reference}
}

\end{document}

\title{\vspace*{7cm}Supplementary Material:
	\ \\
	\ \\
    Demystify Transformers \& Convolutions in Modern Image Deep Networks}

\author{Xiaowei Hu, Min Shi, Weiyun Wang, Sitong Wu, Linjie Xing, Wenhai Wang, Xizhou Zhu, \\ Lewei Lu, Jie Zhou, Xiaogang Wang, Yu Qiao, and Jifeng Dai
	\IEEEcompsocitemizethanks{
		\IEEEcompsocthanksitem Xiaowei Hu is with South China University of Technology, Guangzhou, China.
		\IEEEcompsocthanksitem Min Shi, Weiyun Wang, Sitong Wu, Linjie Xing, Wenhai Wang, and Yu Qiao are with the Shanghai Artificial Intelligence Laboratory, Shanghai, China.
		\IEEEcompsocthanksitem Min Shi is also with Huazhong University of Science and Technology, Wuhan, China, Weiyun Wang is also with Fudan University, Shanghai, China, Sitong Wu is also with The Chinese University of Hong Kong, Shatin, Hong Kong.
		\IEEEcompsocthanksitem Xizhou Zhu, Jie Zhou, and Jifeng Dai are with Tsinghua University, Beijing, China.
		\IEEEcompsocthanksitem Lewei Lu is with SenseTime Research, Beijing, China.
		\IEEEcompsocthanksitem Xiaogang Wang is with The Chinese University of Hong Kong, Shatin, Hong Kong.
		\IEEEcompsocthanksitem Joint first authors: Xiaowei Hu, Min Shi, Weiyun Wang, and Sitong Wu.
		\IEEEcompsocthanksitem Corresponding author: Jifeng Dai (daijifeng@tsinghua.edu.cn).
	}
}

\onecolumn
\maketitle

\markboth{IEEE Transactions on Pattern Analysis and Machine Intelligence}%
{Shell \MakeLowercase{\textit{et al.}}: Bare Demo of IEEEtran.cls for Computer Society Journals}

\maketitle




%%%%%%%%% BODY TEXT
\onecolumn

\vspace*{2mm}
\large
\noindent
There are four parts in this supplementary material.

\begin{itemize}

\vspace*{3mm}
\item
{\bf Part 1} presents the implementation details of the compared STMs and the instantiations of different models.

\vspace*{3mm}
\item
{\bf Part 2} presents more results on ERF analysis.

\vspace*{3mm}
\item
{\bf Part 3} presents more results of invariance analysis.

\vspace*{3mm}
\item
{\bf Part 4} presents more results of robustness analysis.

\end{itemize}

%%%%%%%%%%%%%%%%%%%%%%%%%%%%%%%%%%%%%%%%%%%%%%%%%%%%%%%%%%%%%%%%%%%%%%%%%%%%%%%%%%%%%%%%
\newpage

\clearpage
\vspace*{2mm}
\large
\noindent
{\bf Part 1: Implementation Details of the Compared STMs}
\vspace{10mm}

Table~\ref{stab:model-configuration} specifies the detailed configurations of various models on four scales. These models are instantiated by integrating the STMs into the unified architecture proposed in this study.
Given the diverse complexities of the different STMs, we primarily adhere to the configurations outlined in their respective original papers. When necessary, we make adjustments to the block numbers to ensure alignment of model parameters.

\begin{table}[htbp]
	\newcolumntype{M}[1]{>{\centering\arraybackslash}m{#1}}
	\centering
	\setlength{\tabcolsep}{5pt}
	\caption{The configurations of models with different STMs. The format ``a-b-c-d'' denotes the settings from the first stage to the fourth stage.
	}
	\begin{tabular}{c|c|M{2.7cm}|M{2.7cm}|M{2.7cm}|M{2.7cm}}
		\toprule
		Method &  Component   &  Micro  & Tiny  &  Small  & Base \\
			\midrule
			\multirow{4}{*}{U-HaloNet}  & Block   & 2-2-9-2  & 2-2-6-2  & 2-2-18-2  & 2-2-18-2 \\ 
			& Channel   & 32-64-128-256    & 96-192-384-768     & 96-192-384-768   & 128-256-512-1024 \\
			& Head      & 1-2-4-8          & 3-6-12-24          & 3-6-12-24        & 4-8-16-32 \\
			& MLP Ratio & 4-4-4-4          & 4-4-4-4            & 4-4-4-4          & 4-4-4-4 \\
			\midrule
			\multirow{4}{*}{U-PVT}      & Block   & 2-2-3-2  & 3-4-9-3  & 3-4-21-3  & 3-8-45-3 \\ 
			& Channel   & 32-64-160-256    & 64-128-320-512     & 64-128-320-512   & 64-128-320-512      \\
			& Head      & 1-2-5-8          & 1-2-5-8            & 1-2-5-8          & 1-2-5-8 \\
			& MLP Ratio & 8-8-4-4          & 8-8-4-4            & 8-8-4-4          & 4-4-4-4 \\
			\midrule
			\multirow{4}{*}{U-Swin Transformer}  & Block   & 2-2-9-2  & 2-2-6-2  & 2-2-18-2  & 2-2-18-2 \\ 
			& Channel   & 32-64-128-256    & 96-192-384-768     & 96-192-384-768   & 128-256-512-1024 \\
			& Head      & 1-2-4-8          & 3-6-12-24          & 3-6-12-24        & 4-8-16-32 \\
			& MLP Ratio & 4-4-4-4          & 4-4-4-4            & 4-4-4-4          & 4-4-4-4 \\
			\midrule
			\multirow{3}{*}{U-ConvNeXt} & Block   & 3-3-8-3  & 2-2-9-2  & 2-2-24-2  & 2-2-24-2 \\ 
			& Channel   & 32-64-128-256    & 96-192-384-768     & 96-192-384-768   & 128-256-512-1024 \\
			& MLP Ratio & 4-4-4-4          & 4-4-4-4            & 4-4-4-4          & 4-4-4-4 \\
			\midrule
			\multirow{4}{*}{U-InternImage}  & Block   & 2-2-9-2  & 4-4-18-4  & 4-4-21-4  & 4-4-21-4 \\ 
			& Channel   & 32-64-128-256    & 64-128-256-512     & 80-160-320-640   & 112-224-448-896 \\
			& Head      & 2-4-8-16         & 4-8-16-32          & 5-10-20-40       & 7-14-28-56 \\
			& MLP Ratio & 4-4-4-4          & 4-4-4-4            & 4-4-4-4          & 4-4-4-4 \\
			\bottomrule
		\end{tabular}
		\label{stab:model-configuration} 
	\end{table}
	
	\vspace{5mm}

%	\clearpage

We utilize the official STM implementation for depth-wise convolution, swin attention, spatial reduction attention, and deformable convolution v3. In the absence of an official implementation, we independently implement halo attention. The specific settings tailored for STM are detailed in Table~\ref{tb:stm-configuration}.
	
	\begin{table}[htbp]
		\centering
		\setlength{\tabcolsep}{15pt}
		\caption{Specific settings of STMs. The format ``a-b-c-d'' represents the settings from the first stage to the fourth stage.
		}
		\begin{tabular}{c|c|cc}
			\toprule
			Model                    & STM                              & \multicolumn{2}{c}{Setting}       \\ \midrule
			\rowcolor[HTML]{EFEFEF} 
			\cellcolor[HTML]{EFEFEF} & \cellcolor[HTML]{EFEFEF}         & Window Size             & 7-7-7-7 \\
			\rowcolor[HTML]{EFEFEF} 
			\multirow{-2}{*}{\cellcolor[HTML]{EFEFEF}U-HaloNet} & \multirow{-2}{*}{\cellcolor[HTML]{EFEFEF}Halo Attention} & Halo Width & 3-3-3-3 \\
			\rowcolor[HTML]{FFFFFF} 
			U-PVT                    & Spatial Reduction Attention & Spatial Reduction Ratio & 8-4-2-1 \\
			\rowcolor[HTML]{EFEFEF} 
			U-Swin Transformer       & Shifted Window Attention         & Window Size             & 7-7-7-7 \\
			\rowcolor[HTML]{FFFFFF} 
			U-ConvNeXt               & Depth-Wise Convolution           & Kernel Size             & 7-7-7-7 \\
			\rowcolor[HTML]{EFEFEF} 
			U-InternImage           & Deformable Convolution v3        & Sampling Points         & 9-9-9-9 \\ \bottomrule
		\end{tabular}
		\label{tb:stm-configuration} 
	\end{table}

\clearpage

%%%%%%%%%%%%%%%%%%%%%%%%%%%%%%%%%%%%%%%%%%%%%%%%%%%%%%%%%%%%%%%%%%%%%%%%%%%%%%%%%%%%%%%%
\newpage

\clearpage
\vspace*{2mm}
\large
\noindent
{\bf Part 2: More Results of ERF Analysis}
\vspace{10mm}

\begin{figure}[htbp]
	\centering
	\includegraphics[width=0.8\linewidth]{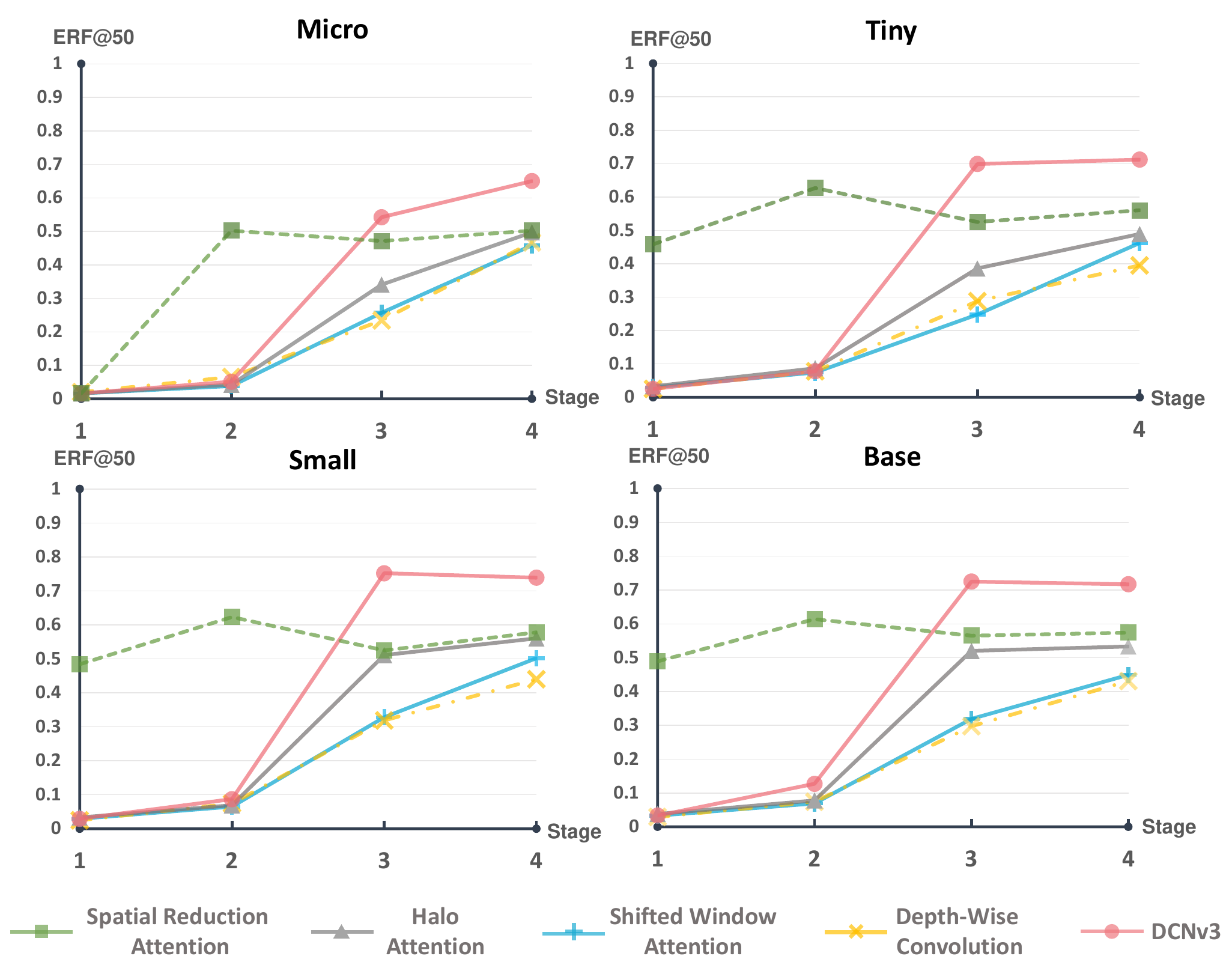}
	\caption{Effective Receptive Fields (ERFs) of Various STMs: The ERF@50 curves illustrate the comparative analysis of different STMs across four model scales.}
	\label{fig:erf-comparison-curve11111}
\end{figure}

Here, we present additional quantitative results for the ERF (Effective Receptive Field) analysis. In Figure~\ref{fig:erf-comparison-curve11111}, we illustrate the ERF@50 curves for all compared models, employing various scales.
Upon examining the figure, the following observations emerge:
(i) In PVT with global attention and no locality inductive bias, we observe a significantly larger ERF during the first two stages compared to other STMs. However, its ERF becomes comparable to other STMs in the last two stages when handling inputs with a resolution of $224\times 224$.
(ii) Swin attention and the $7\times 7$ depth-wise convolution in ConvNeXt exhibit similar ERF sizes.
(iii) Notably, halo attention and DCNv3 demonstrate a steep increase in ERF during the last two stages, with DCNv3 achieving the largest ERF among all STMs.
These findings suggest that the dynamic sparse sampling strategy effectively learns to capture long-range dependencies, proving more effective than simply enlarging the sampling window, as observed in the case of halo attention.

\clearpage

Here, we present a more comprehensive set of ERF quantitative results in Table~\ref{stab:erf-results}, incorporating ERF@20, ERF@50, and ERF@99 as the evaluation metrics. %Please refer to line 692 in the main paper for the definition of these metrics.

\vspace{5mm}

\begin{table}[h]
	\centering
	\setlength\tabcolsep{3pt}
	\caption{ERF Results for Various STMs.}
	\label{stab:erf-results}
	\resizebox{0.99\linewidth}{!}{
		\begin{tabular}{@{}l|c|rrrr|rrrr|rrrr|rrrr@{}}
			\toprule
			& Scale  & \multicolumn{4}{c|}{Micro} & \multicolumn{4}{c|}{Tiny} & \multicolumn{4}{c|}{Small} & \multicolumn{4}{c}{Base}  \\ \midrule
			& Stage  & 1     & 2    & 3    & 4    & 1    & 2    & 3    & 4    & 1     & 2    & 3    & 4    & 1    & 2    & 3    & 4    \\ \midrule
			\multirow{3}{*}{U-HaloNet}     & ERF@20 & 1.1   & 2.0  & 10.9 & 18.1 & 1.1  & 3.3  & 12.7 & 18.1 & 1.1   & 2.9  & 21.2 & 23.9 & 1.1  & 2.9  & 23.9 & 23.0 \\
			& ERF@50 & 1.6   & 4.2  & 34.1 & 49.7 & 3.3  & 8.7  & 38.6 & 48.9 & 3.3   & 6.9  & 51.1 & 56.0 & 3.8  & 7.8  & 52.0 & 53.3 \\
			& ERF@99 & 17.2  & 34.6 & 98.0 & 98.9 & 21.2 & 41.7 & 98.4 & 98.9 & 21.2  & 40.0 & 98.4 & 98.9 & 21.2 & 41.7 & 98.4 & 98.9 \\ \midrule
			\multirow{3}{*}{U-PVT}         & ERF@20 & 1.1   & 6.9  & 7.9  & 12.3 & 1.6  & 23.0 & 14.1 & 16.7 & 1.6   & 20.3 & 13.2 & 16.7 & 1.6  & 20.8 & 15.8 & 15.4 \\
			& ERF@50 & 1.6   & 50.2 & 47.1 & 50.2 & 45.8 & 62.7 & 52.5 & 56.0 & 48.4  & 62.3 & 52.5 & 57.8 & 48.9 & 61.4 & 56.5 & 57.4 \\
			& ERF@99 & 98.0  & 99.3 & 98.9 & 99.3 & 98.9 & 99.3 & 99.3 & 99.3 & 98.9  & 99.3 & 99.3 & 99.3 & 98.9 & 99.3 & 99.3 & 99.3 \\ \midrule
			\multirow{3}{*}{U-Swin}        & ERF@20 & 0.7   & 2.0  & 7.8  & 17.2 & 1.1  & 2.9  & 9.6  & 18.1 & 1.1   & 2.5  & 14.5 & 23.4 & 1.1  & 2.9  & 14.1 & 19.0 \\
			& ERF@50 & 1.6   & 3.8  & 25.7 & 45.8 & 2.9  & 7.4  & 24.8 & 46.2 & 2.9   & 6.5  & 32.8 & 50.2 & 3.3  & 6.9  & 31.9 & 44.9 \\
			& ERF@99 & 10.0  & 24.3 & 94.9 & 98.4 & 12.3 & 28.3 & 94.4 & 98.8 & 12.3  & 27.5 & 95.8 & 98.9 & 12.3 & 27.9 & 96.2 & 99.8 \\ \midrule
			\multirow{3}{*}{U-ConvNeXt}    & ERF@20 & 1.1   & 2.9  & 10.5 & 22.5 & 1.1  & 3.8  & 10.9 & 16.3 & 1.1   & 3.3  & 15.0 & 20.8 & 1.1  & 3.3  & 13.6 & 19.9 \\
			& ERF@50 & 2.0   & 6.5  & 23.4 & 46.7 & 2.5  & 7.8  & 28.8 & 39.5 & 2.5   & 7.4  & 31.9 & 44.0 & 2.9  & 7.4  & 29.7 & 43.1 \\
			& ERF@99 & 9.2   & 26.1 & 87.7 & 98.9 & 10.5 & 23.9 & 97.1 & 98.4 & 10.0  & 23.4 & 98.0 & 98.4 & 10.5 & 23.9 & 97.5 & 98.4 \\ \midrule
			\multirow{3}{*}{U-InternImage} & ERF@20 & 0.7   & 2.5  & 15.0 & 25.7 & 1.1  & 3.3  & 27.0 & 35.9 & 1.1   & 3.8  & 34.1 & 36.4 & 1.6  & 5.1  & 37.7 & 39.1 \\
			& ERF@50 & 1.6   & 5.1  & 54.2 & 65.0 & 2.5  & 7.8  & 69.9 & 71.2 & 2.9   & 8.7  & 75.2 & 73.9 & 3.3  & 12.7 & 72.5 & 71.7 \\
			& ERF@99 & 7.4   & 30.6 & 99.3 & 99.3 & 19.0 & 41.3 & 99.8 & 99.8 & 20.8  & 45.3 & 99.8 & 99.8 & 27.0 & 71.2 & 99.8 & 99.8 \\ \bottomrule
		\end{tabular}
	}
\end{table}

\clearpage

We depict the ERF maps for micro, tiny, small, and base models in the subsequent figures.
\begin{figure}[htbp]
	\centering
	\includegraphics[width=0.7\linewidth]{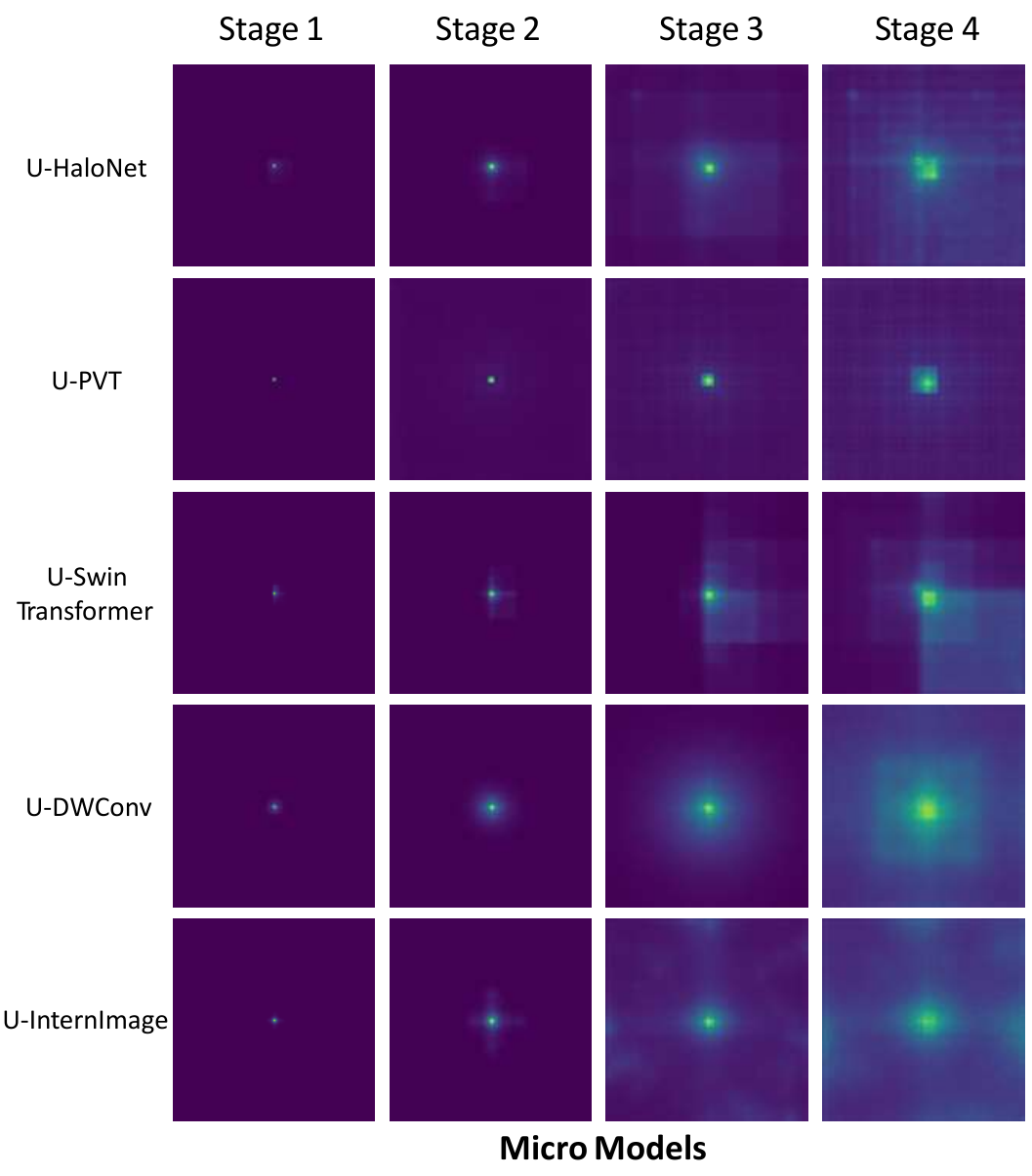}
	\caption{Visualization of ERF for Micro Models.}
	\label{fig:erf-comparison-curve}
\end{figure}
\begin{figure}[htbp]
	\centering
	\includegraphics[width=0.7\linewidth]{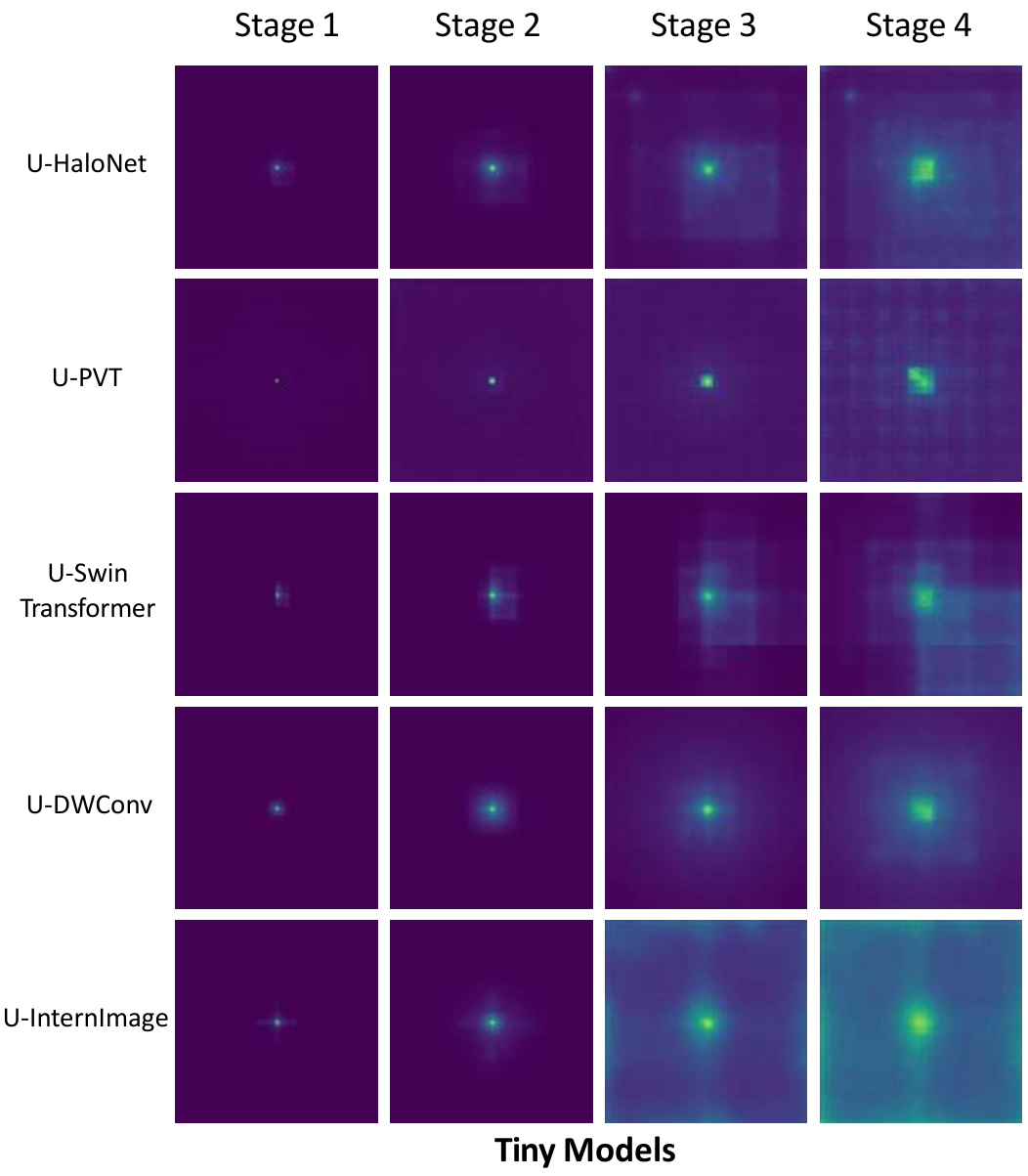}
	\caption{Visualization of ERF for Tiny Models.}
	\label{fig:erf-comparison-curve}
\end{figure}
\begin{figure}[htbp]
	\centering
	\includegraphics[width=0.7\linewidth]{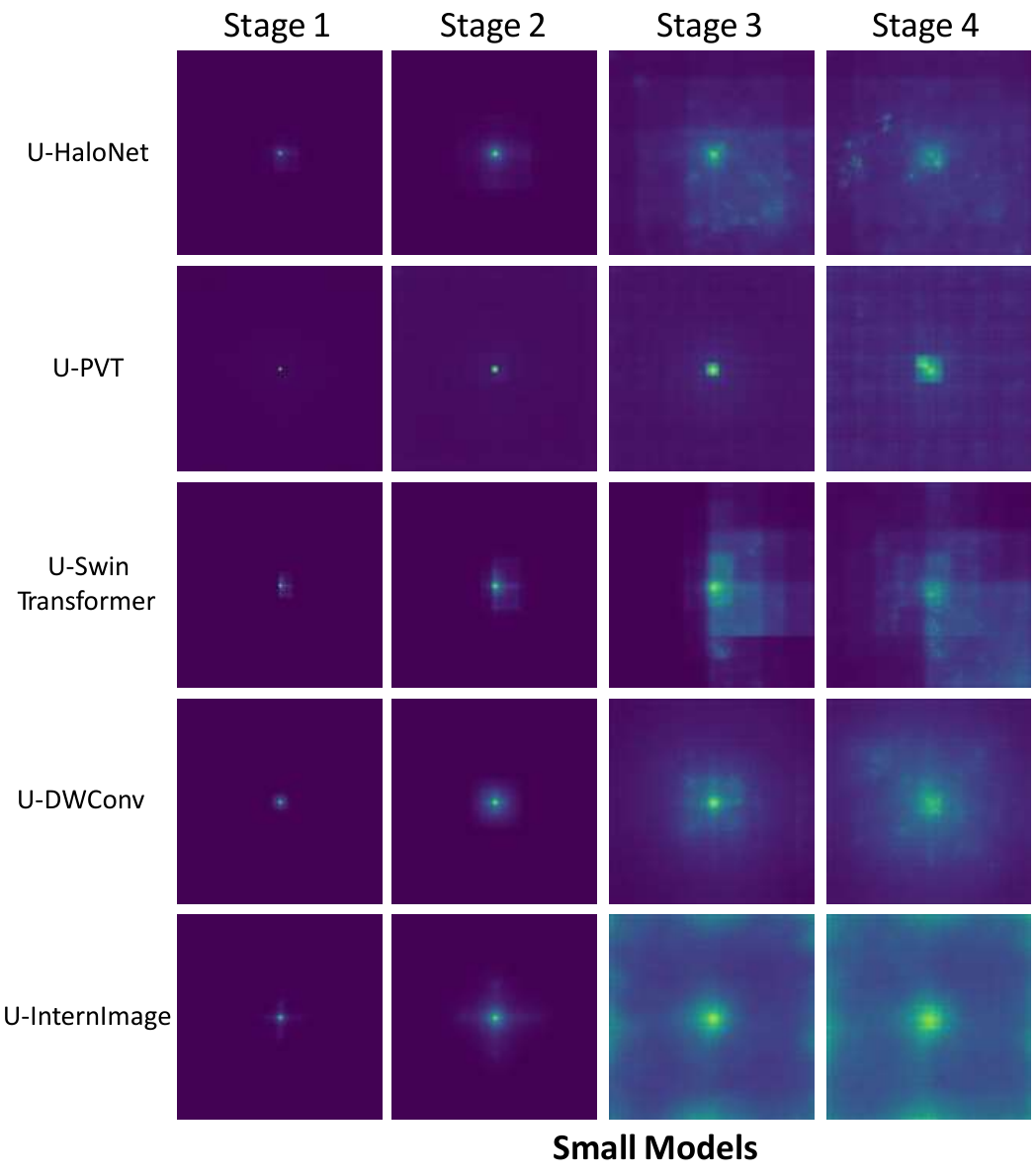}
	\caption{Visualization of ERF for Small Models.}
	\label{fig:erf-comparison-curve}
\end{figure}
\begin{figure}[htbp]
	\centering
	\includegraphics[width=0.7\linewidth]{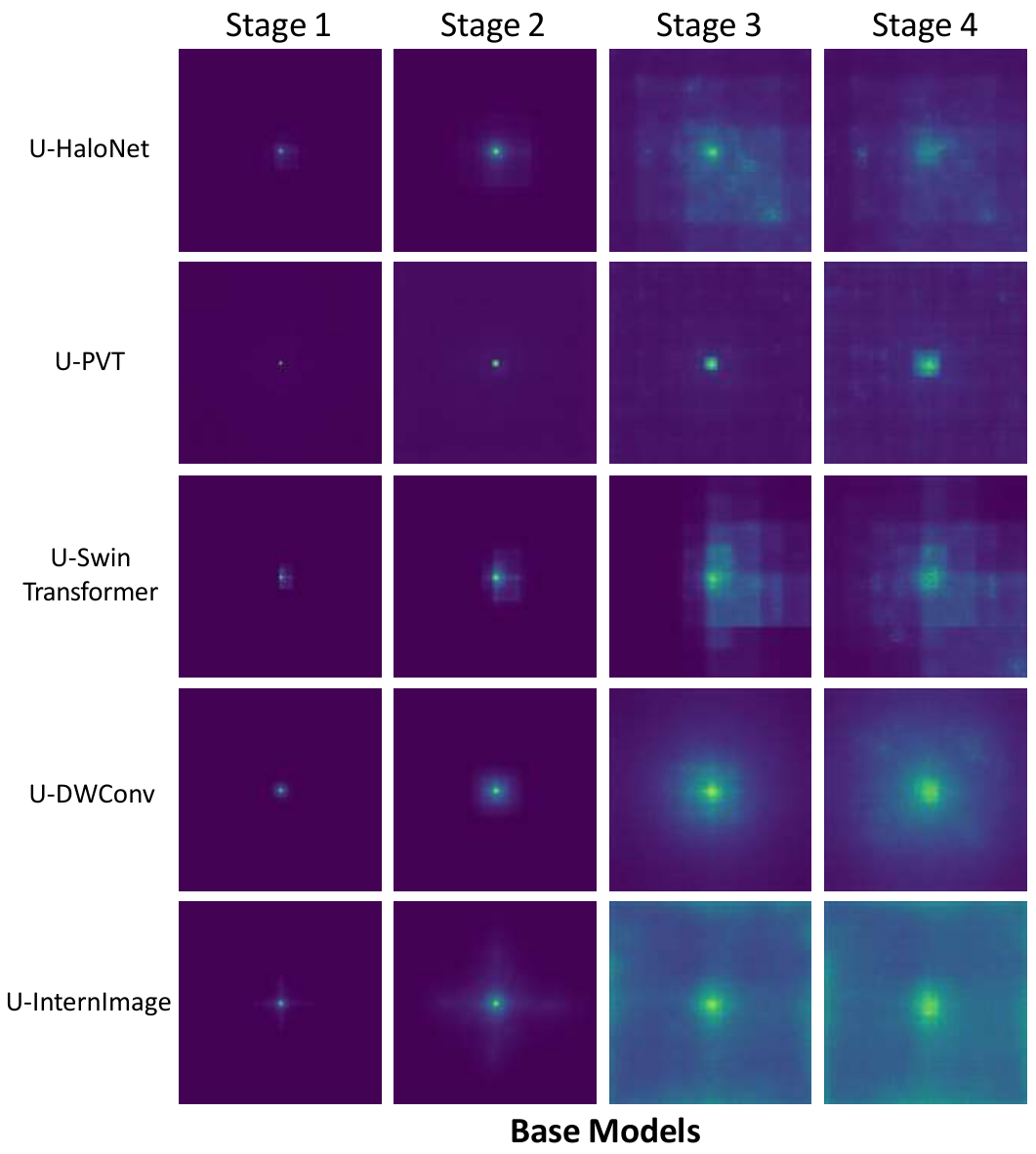}
	\caption{Visualization of ERF for Base Models.}
	\label{fig:erf-comparison-curve}
\end{figure}

\clearpage

%%%%%%%%%%%%%%%%%%%%%%%%%%%%%%%%%%%%%%%%%%%%%%%%%%%%%%%%%%%%%%%%%%%%%%%%%%%%%%%%%%%%%%%%

\newpage
\vspace*{2mm}
\large
\noindent
{\bf Part 3: More Results of Invariance Analysis}
\vspace{10mm}

In the main manuscript, we evaluate the invariance of various STMs with input transformations. More detailed results are presented in Table~\ref{stab:translation-invariance}, Table~\ref{stab:rotation-invariance} and Table~\ref{stab:scaling-invariance}. 

\vspace{10mm}

\begin{table}[h]
	\centering
	\caption{{Results of translation invariance on ImageNet-1k.} ``Con.'' and ``Acc.'' denote ``Consistency'' and ``Accuracy'', respectively.}
	\label{stab:translation-invariance}
	\begin{tabular}{c|cc|cc|cc|cc|cc}
		\toprule
		Translation & \multicolumn{2}{c|}{U-HaloNet} & \multicolumn{2}{c|}{U-PVT} & \multicolumn{2}{c|}{U-Swin} & \multicolumn{2}{c|}{U-ConvNeXt} & \multicolumn{2}{c}{U-InternImage} \\
		(pixels)  & Con.            & Acc.           & Con.          & Acc.         & Con.                & Acc.               & Con.            & Acc.            & Con.             & Acc.             \\ \midrule
		0                             & 99.7           & 83.0          & 99.7         & 82.1        & 99.7               & 82.3              & 99.8           & 82.2           & 99.8            & 83.3            \\
		4                             & 95.8           & 83.0          & 95.2         & 82.1        & 95.5               & 82.3              & 96.3           & 82.2           & 96.5            & 83.2            \\
		8                             & 95.1           & 82.9          & 94.4         & 81.9        & 94.8               & 82.2              & 95.7           & 82.1           & 95.9            & 83.2            \\
		12                            & 94.5           & 82.8          & 93.6         & 81.9        & 94.1               & 82.1              & 95.2           & 82.0           & 95.3            & 83.2            \\
		16                            & 95.1           & 82.9          & 93.7         & 81.9        & 94.3               & 82.2              & 95.7           & 82.1           & 95.7            & 83.0            \\
		20                            & 93.6           & 82.7          & 92.6         & 81.8        & 93.2               & 82.0              & 94.3           & 81.9           & 94.4            & 82.9            \\
		24                            & 93.3           & 82.7          & 92.4         & 81.7        & 92.8               & 81.9              & 93.8           & 81.7           & 93.9            & 82.8            \\
		28                            & 92.8           & 82.4          & 92.1         & 81.6        & 92.2               & 81.6              & 93.2           & 81.5           & 93.4            & 82.5            \\
		32                            & 92.9           & 82.3          & 92.1         & 81.5        & 92.0               & 81.5              & 93.3           & 81.5           & 93.4            & 82.4            \\
		36                            & 92.1           & 82.2          & 91.2         & 81.1        & 91.5               & 81.5              & 92.8           & 81.5           & 92.6            & 82.3            \\
		40                            & 91.7           & 82.0          & 90.8         & 81.0        & 91.1               & 81.2              & 92.0           & 81.0           & 92.4            & 82.1            \\
		44                            & 91.2           & 81.7          & 90.4         & 80.7        & 90.5               & 80.9              & 91.7           & 81.0           & 91.9            & 81.9            \\
		48                            & 91.2           & 81.4          & 90.4         & 80.6        & 90.2               & 80.6              & 91.4           & 80.7           & 91.9            & 81.7            \\
		52                            & 90.2           & 81.1          & 89.3         & 80.2        & 89.5               & 80.3              & 90.3           & 80.3           & 90.9            & 81.3            \\
		56                            & 89.9           & 80.9          & 88.8         & 80.0        & 88.8               & 79.9              & 89.7           & 80.1           & 90.3            & 81.2            \\
		60                            & 88.9           & 80.4          & 88.1         & 79.6        & 88.2               & 79.6              & 88.9           & 79.5           & 89.4            & 80.6            \\
		64                            & 88.6           & 80.1          & 87.6         & 79.2        & 87.8               & 79.3              & 88.5           & 79.0           & 89.2            & 80.6            \\ \bottomrule
	\end{tabular}
\end{table}

\vspace{10mm}

\begin{table}[h]
	\newcommand{\tabincell}[2]{\begin{tabular}{@{}#1@{}}#2\end{tabular}}
	\centering
	\caption{{Results of rotation invariance on ImageNet-1k.} ``Con.'' and ``Acc.'' denote ``Consistency'' and ``Accuracy'', respectively.}
	\label{stab:rotation-invariance}
	\begin{tabular}{c|cc|cc|cc|cc|cc}
		\hline
		\multirow{2}{*}{\tabincell{c}{Rotation \\ (degree)}}         & \multicolumn{2}{c|}{U-HaloNet}     & \multicolumn{2}{c|}{U-PVT}         & \multicolumn{2}{c|}{U-Swin}        & \multicolumn{2}{c|}{U-ConvNeXt}    & \multicolumn{2}{c}{U-InternImage} \\
		& Con.   & Acc.                       & Con.   & Acc.                       & Con.   & Acc.                       & Con.   & Acc.                       & Con.              & Acc.            \\ \hline
		\multicolumn{1}{c|}{0}       & 100.0 & \multicolumn{1}{c|}{83.0} & 100.0 & \multicolumn{1}{c|}{82.1} & 100.0 & \multicolumn{1}{c|}{82.3} & 100.0 & \multicolumn{1}{c|}{82.2} & 100.0            & 83.3           \\
		\multicolumn{1}{c|}{5}       & 93.2  & \multicolumn{1}{c|}{82.4} & 92.9  & \multicolumn{1}{c|}{81.5} & 93.1  & \multicolumn{1}{c|}{81.7} & 93.4  & \multicolumn{1}{c|}{81.4} & 93.3             & 82.3           \\
		\multicolumn{1}{c|}{10}      & 93.2  & \multicolumn{1}{c|}{82.4} & 92.4  & \multicolumn{1}{c|}{81.4} & 92.7  & \multicolumn{1}{c|}{81.6} & 93.3  & \multicolumn{1}{c|}{81.4} & 94.0             & 82.7           \\
		\multicolumn{1}{c|}{15}      & 93.1  & \multicolumn{1}{c|}{82.3} & 92.3  & \multicolumn{1}{c|}{81.4} & 92.4  & \multicolumn{1}{c|}{81.5} & 93.1  & \multicolumn{1}{c|}{81.2} & 93.7             & 82.5           \\
		\multicolumn{1}{c|}{20}      & 92.6  & \multicolumn{1}{c|}{82.1} & 92.2  & \multicolumn{1}{c|}{81.2} & 92.0  & \multicolumn{1}{c|}{81.3} & 92.5  & \multicolumn{1}{c|}{81.0} & 93.4             & 82.3           \\
		\multicolumn{1}{c|}{25}      & 92.2  & \multicolumn{1}{c|}{81.8} & 91.6  & \multicolumn{1}{c|}{80.9} & 91.5  & \multicolumn{1}{c|}{81.0} & 91.8  & \multicolumn{1}{c|}{80.6} & 93.0             & 82.1           \\
		\multicolumn{1}{c|}{30}      & 91.6  & \multicolumn{1}{c|}{81.6} & 91.4  & \multicolumn{1}{c|}{80.9} & 90.9  & \multicolumn{1}{c|}{80.7} & 90.8  & \multicolumn{1}{c|}{80.2} & 92.6             & 81.9           \\
		\multicolumn{1}{c|}{35}      & 90.6  & \multicolumn{1}{c|}{81.1} & 90.2  & \multicolumn{1}{c|}{80.2} & 89.6  & \multicolumn{1}{c|}{80.0} & 89.7  & \multicolumn{1}{c|}{79.5} & 91.7             & 81.5           \\
		\multicolumn{1}{c|}{40}      & 88.2  & \multicolumn{1}{c|}{79.6} & 87.0  & \multicolumn{1}{c|}{78.5} & 87.0  & \multicolumn{1}{c|}{78.4} & 86.6  & \multicolumn{1}{c|}{77.8} & 90.4             & 80.9           \\
		\multicolumn{1}{c|}{45}      & 85.4  & \multicolumn{1}{c|}{77.7} & 83.6  & \multicolumn{1}{c|}{76.2} & 84.2  & \multicolumn{1}{c|}{76.4} & 83.5  & \multicolumn{1}{c|}{75.7} & 86.2             & 78.3           \\ \hline
	\end{tabular}
\end{table}

\begin{table}[tp]
	\newcommand{\tabincell}[2]{\begin{tabular}{@{}#1@{}}#2\end{tabular}}
	\centering
	\caption{{Results of scaling invariance on COCO.} ``Con.'' and ``mAP'' denote ``Consistency'' and ``box mAP'' respectively.}
	\label{stab:scaling-invariance}
	\begin{tabular}{c|cc|cc|cc|cc|cc}
		\toprule
		\multirow{2}{*}{\tabincell{c}{Scale \\ Factor}}
		& \multicolumn{2}{c|}{U-HaloNet} & \multicolumn{2}{c|}{U-PVT} & \multicolumn{2}{c|}{U-Swin} & \multicolumn{2}{c|}{U-ConvNeXt} & \multicolumn{2}{c}{U-InternImage} \\
		& Con.            & mAP           & Con.          & mAP         & Con.           & mAP         & Con.             & mAP           & Con.              & mAP            \\ \midrule
		0.25     & 14.6           & 24.8          & 8.9          & 16.4        & 13.3          & 23.3        & 13.5            & 23.7          & 15.0             & 25.2           \\
		0.50     & 29.1           & 39.2          & 25.0         & 34.8        & 26.4          & 37.0        & 27.5            & 37.4          & 29.9             & 39.1           \\
		0.75     & 40.2           & 44.8          & 36.9         & 41.6        & 35.8          & 42.3        & 40.3            & 42.6          & 42.7             & 45.0           \\
		1.00     & 100.0          & 46.9          & 100.0        & 44.2        & 100.0         & 44.3        & 100.0           & 44.3          & 100.0            & 47.2           \\
		1.25     & 46.2           & 46.9          & 42.8         & 44.7        & 41.1          & 44.1        & 46.8            & 44.2          & 49.7             & 47.5           \\
		1.50     & 46.2           & 45.4          & 38.7         & 43.1        & 37.2          & 42.3        & 39.9            & 42.3          & 43.5             & 46.3           \\
		1.75     & 37.4           & 42.7          & 34.9         & 40.0        & 33.1          & 39.4        & 34.8            & 39.5          & 39.5             & 44.2           \\
		2.00     & 33.9           & 39.8          & 31.1         & 35.8        & 30.0          & 36.1        & 30.6            & 36.2          & 35.9             & 41.7           \\
		2.25     & 30.1           & 36.6          & 26.6         & 31.6        & 26.3          & 33.0        & 26.7            & 33.0          & 32.3             & 38.9           \\
		2.50     & 26.9           & 33.5          & 22.8         & 27.8        & 23.3          & 30.1        & 23.6            & 30.0          & 29.3             & 36.2           \\
		2.75     & 24.1           & 30.9          & 19.4         & 24.2        & 20.8          & 27.4        & 21.0            & 27.1          & 26.5             & 33.4           \\
		3.00     & 21.6           & 27.8          & 15.9         & 20.2        & 18.3          & 24.6        & 18.6            & 24.5          & 24.1             & 30.9           \\ \bottomrule
	\end{tabular}
\end{table}
\clearpage
%%%%%%%%%%%%%%%%%%%%%%%%%%%%%%%%%%%%%%%%%%%%%%%%%%%%%%%%%%%%%%%%%%%%%%%%%%%%%%%%%%%%%%%%

%%%%%%%%%%%%%%%%%%%%%%%%%%%%%%%%%%%%%%%%%%%%%%%%%%%%%%%%%%%%%%%%%%%%%%%%%%%%%%%%%%%%%%%%

\newpage
\vspace*{2mm}
\large
\noindent
{\bf Part 4: More Results of Robustness Analysis}
\vspace{10mm}

\begin{table}[h]
\centering
\setlength\tabcolsep{3pt}
	\caption{Evaluation of model performance on ImageNet-1k under various natural noise conditions and strengths of adversarial attacks.}
	\label{stab:erf-results}
	\resizebox{0.99\linewidth}{!}{
\begin{tabular}{c|c|c|c|c|c|c|c|c|c}
\hline
Scale & Method & Clean Acc & ImageNet-A Accuracy & ImageNet-O AUPR & ImageNet-P NmFP & ImageNet-C 1-mCE & FGSM Adv ($\epsilon=8/255$) & FGSM Adv ($\epsilon=2/255$) & FGSM Adv ($\epsilon=0.5/255$) \\ \cline{1-10}
\multirow{5}{*}{Micro} & U-HaloNet & 75.8 & 9.3 & 17.3 & -0.066 & 46.3 & 21.4 & 24.7 & 36.7 \\ \cline{2-10}
 & U-PVT & 72.8 & 5.4 & 16.9 & -0.073 & 40.4 & 14.5 & 16.4 & 28 \\ \cline{2-10}
 & U-Swin Transformer & 74.4 & 7.2 & 16.8 & -0.072 & 42.6 & 18.9 & 21.3 & 32 \\ \cline{2-10}
 & U-DWConv & 75.1 & 5.7 & 16.4 & -0.064 & 43.3 & 18.6 & 20.3 & 32.3 \\ \cline{2-10}
 & U-InternImage & 75.3 & 6.8 & 16.8 & -0.063 & 44.8 & 17.2 & 19.5 & 32.6 \\ \hline
\multirow{5}{*}{Tiny} & U-HaloNet & 83.0 & 30.7 & 22.6 & -0.038 & 57.9 & 38 & 40.7 & 51.3 \\ \cline{2-10}
 & U-PVT & 82.1 & 26.1 & 21.8 & -0.044 & 56.1 & 41.2 & 43.5 & 51.5 \\ \cline{2-10}
 & U-Swin Transformer & 82.3 & 27.0 & 21.9 & -0.041 & 55.5 & 32.7 & 36.1 & 47.8 \\ \cline{2-10}
 & U-DWConv & 82.2 & 23.3 & 20.7 & -0.038 & 55.4 & 32.3 & 36 & 50.2 \\ \cline{2-10}
 & U-InternImage & 83.3 & 31.9 & 22.7 & -0.037 & 58.9 & 41.7 & 45.4 & 55.6 \\ \hline
\multirow{5}{*}{Small} & U-HaloNet & 84.0 & 37.8 & 25.2 & -0.036 & 61.2 & 42.5 & 46.2 & 57.3 \\ \cline{2-10}
 & U-PVT & 83.2 & 33.8 & 21.8 & -0.038 & 58.4 & 41.4 & 44.1 & 53.3 \\ \cline{2-10}
 & U-Swin Transformer & 83.3 & 33.1 & 23.8 & -0.039 & 58.4 & 39.2 & 43.6 & 54.9 \\ \cline{2-10}
 & U-DWConv & 83.1 & 29.2 & 21.8 & -0.037 & 58.1 & 37.5 & 41.3 & 52.7 \\ \cline{2-10}
 & U-InternImage & 84.1 & 38.0 & 24.1 & -0.033 & 60.9 & 40 & 44.2 & 56.5 \\ \hline
\multirow{5}{*}{Base} & U-HaloNet & 84.6 & 41.6 & 26.1 & -0.027 & 62.0 & 46.5 & 49.6 & 59.2 \\ \cline{2-10}
 & U-PVT & 83.4 & 35.5 & 23.5 & -0.038 & 59.1 & 41.5 & 45.1 & 55.3 \\ \cline{2-10}
 & U-Swin Transformer & 83.7 & 36.4 & 24.2 & -0.036 & 59.0 & 41.6 & 45.3 & 56.6 \\ \cline{2-10}
 & U-DWConv & 83.7 & 34.4 & 21.1 & -0.034 & 59.0 & 42.4 & 46.4 & 58 \\ \cline{2-10}
 & U-InternImage & 84.5 & 42.0 & 26.1 & -0.033 & 61.7 & 38.5 & 42.9 & 56.3 \\ \hline
\end{tabular} }
\label{tab:performance_metrics}
\end{table}

To perform an evaluation of natural noise and adversarial robustness, each model (U-HaloNet, U-PVT, U-Swin Transformer, U-DWConv, U-InternImage) is trained on the ImageNet dataset. 
The robustness of natural noise is evaluated using ImageNet-A accuracy, ImageNet-O AUPR, ImageNet-P NmFP, and ImageNet-C 1-mCE. The results reveal that U-HaloNet and U-InternImage consistently demonstrate superior performance across different scales (Micro, Tiny, Small, Base). These models excel in handling out-of-distribution samples and common corruptions, maintaining high clean accuracy. In contrast, U-PVT performs the worst overall, while U-Swin Transformer and U-DWConv show moderate performance.
For adversarial robustness, adversarial examples are generated using the Fast Gradient Sign Method (FGSM) with perturbation strengths ($\epsilon$) of 8/255, 2/255, and 0.5/255. U-HaloNet achieves the highest clean accuracy and strong performance against adversarial attacks, particularly at higher perturbation strengths. U-InternImage also shows high clean accuracy and robustness to adversarial attacks. Conversely, U-PVT performs poorly, with low robustness against adversarial attacks. U-Swin Transformer and U-DWConv demonstrate moderate robustness, with U-Swin Transformer performing better at lower perturbation strengths. 
Overall, U-HaloNet and U-InternImage stand out as the most robust models, effectively handling both natural noise and adversarial perturbations across various scales.

\clearpage

\ifCLASSOPTIONcaptionsoff
  \newpage
\fi



%\bibliographystyle{IEEEtran}
%\begin{thebibliography}{1}


%% file: Section0-abstract.tex
\begin{abstract}

Vision transformers have gained popularity recently, leading to the development of new vision backbones with improved features and consistent performance gains. However, these advancements are not solely attributable to novel feature transformation designs; certain benefits also arise from advanced network-level and block-level architectures.
This paper aims to identify the real gains of popular convolution and attention operators through a detailed study. We find that the key difference among these feature transformation modules, such as attention or convolution, lies in their spatial feature aggregation approach, known as the ``spatial token mixer'' (STM).
To facilitate an impartial comparison, we introduce a unified architecture to neutralize the impact of divergent network-level and block-level designs. 
Subsequently, various STMs are integrated into this unified framework for comprehensive comparative analysis. 
Our experiments on various tasks and an analysis of inductive bias show a significant performance boost due to advanced network-level and block-level designs, but performance differences persist among different STMs.
Our detailed analysis also reveals various findings about different STMs, including effective receptive fields, invariance, and adversarial robustness tests.
% %

\end{abstract}

%% file: Section1-introduction.tex
\section{Introduction}\label{sec:intro}

Vision transformers~\cite{dosovitskiy2020image} have revolutionized the landscape of visual backbone models, inspiring a set of deep networks incorporating attention~\cite{vaswani2021scaling,liu2021swin,wang2021pyramid}, convolution~\cite{liu2022convnet,ding2022scaling,liu2022more}, and hybrid~\cite{dai2021coatnet,rao2022hornet} blocks.
These networks consistently demonstrate performance gains attributed to novel feature transformation operators. 
However, it is essential to acknowledge the potential influence of advanced network engineering techniques, including macro architecture design~\cite{DBLP:conf/cvpr/YuLZSZWFY22} and training recipes~\cite{liu2022convnet,liu2021swin,pmlr-v139-touvron21a}. 
This broader context prompts a critical question that \emph{do the observed performance gains primarily stem from novel operator designs or from the adoption of advanced network engineering techniques?}

\begin{figure}[tp]
	\centering
	\includegraphics[width=0.99\linewidth]{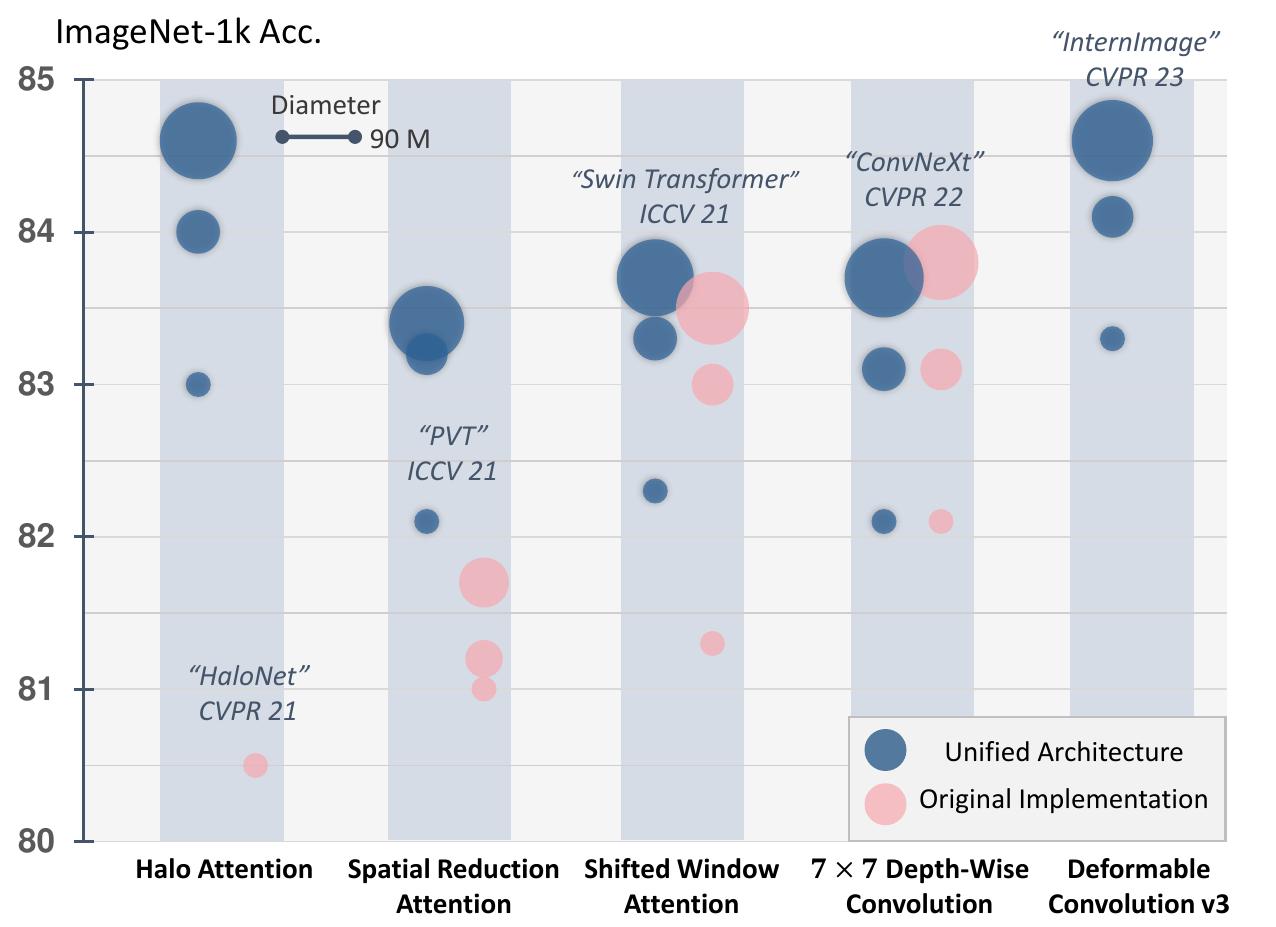}
    \caption{Evolution of spatial token mixers. While consistent performance gains are observed using the original implementation or reported results, unifying network engineering techniques (overall architecture and training recipe) reveals that relatively early STMs, such as halo attention, achieve state-of-the-art performance.}
	\label{fig:teaser}
    \vspace{-3mm}
\end{figure}

\begin{figure*}[tp]
	\includegraphics[width=\linewidth]{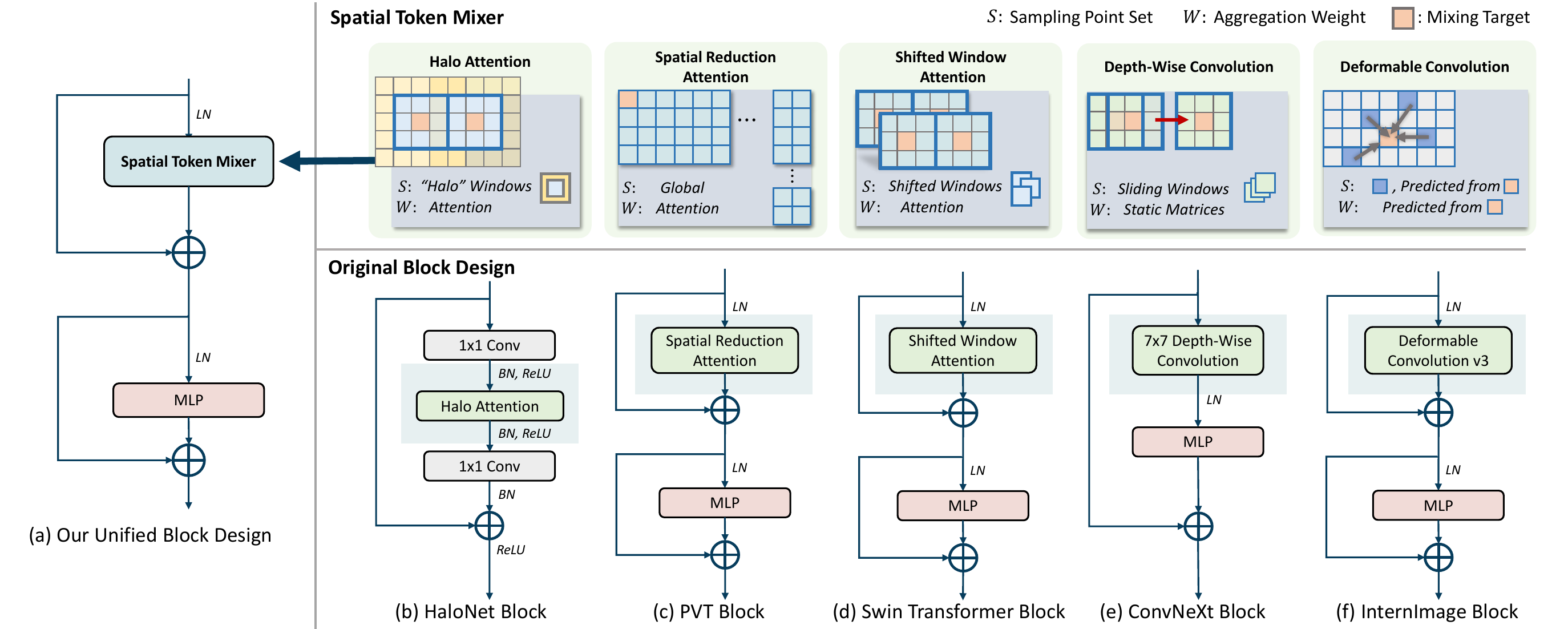}
	\caption{The schematic illustration depicts (a) our unified block design (following the vanilla Transformer~\cite{vaswani2017attention} block design) and spatial token mixer designs (top), along with the original block designs (bottom) of (b) HaloNet, (c) PVT, (d) Swin Transformer, (e) ConvNeXt, and (f) InternImage. }
	\label{fig:arc}
  \vspace{-3mm}
\end{figure*}

To address this inquiry, we conduct a comprehensive survey of recent vision backbones featuring novel operators~\cite{vaswani2021scaling,liu2021swin,wang2021pyramid,liu2022convnet,dong2022cswin}. Our examination reveals that the distinctions among these operators, and their asserted novelty, primarily hinge on the approach to spatial feature aggregation, commonly referred to as the Spatial Token Mixer (STM).
Notably, our scrutiny aligns with the broader interest in understanding the true efficacy of STMs.
A recent study, MetaFormer~\cite{DBLP:conf/cvpr/YuLZSZWFY22}, underscores the significance of transformer block design in overall performance. 
Remarkably, MetaFormer demonstrates that the STM within the transformer block, specifically the attention layer, can be substituted with a simple mixing operation, such as pooling, without a substantial drop in performance. 
This finding serves as an additional motivation for our investigation into the authentic performance gains associated with various STMs, aiming to discover any potential biases arising from disparate network engineering practices.

To achieve this, we first construct a modern and unified architecture for comparison, where we sequentially determine the stage-level design~(\ie, stem and transition layers) and block-level design~(\ie, the topology of basic blocks) by a series of experiments and analyses. 
We find that some macro designs such as different downsampling operations may affect the performances for a certain type of STM, and the proper block design can also contribute to performance gain significantly.
We combine the findings and beneficial designs into a unified architecture. By instantiating different STMs on this architecture, we achieve comparable or even better results than the reported performance in their original papers, as shown in Fig.~\ref{fig:teaser}. 
Our study includes four pivotal STMs representing diverse approaches in vision backbones: local attention~\cite{vaswani2021scaling, liu2021swin}, global attention~\cite{wang2021pyramid}, depth-wise convolution~\cite{liu2022convnet}, and dynamic convolution~\cite{2023intern}. Refer to Fig.~\ref{fig:arc} for a visual overview of these operations. 
Each STM is selected for distinct characteristics, ensuring a thorough exploration of spatial feature aggregation in the unified architecture.

Employing the instantiated models, we conduct an investigation and comparison of STMs across multiple dimensions.
First, we assess the performance of different models on both upstream (classification) and downstream (object detection) tasks, using the ImageNet-1k~\cite{deng2009imagenet} and COCO~\cite{lin2014microsoft} datasets. 
Classification and object detection represent fundamental and widely studied challenges in computer vision, encompassing both global and local vision tasks.
Our findings, as illustrated in Fig.~\ref{fig:teaser}, reveal performance gaps among various STMs, deviating from reported trends with original implementations. Notably, earlier STMs imbued with advanced design principles achieve comparable or superior performance compared to recent counterparts in both upstream and downstream tasks.
Second, STMs exhibiting vision-specific inductive biases and dynamic information modeling mechanisms, such as local attention and dynamic convolution, demonstrate relatively robust performance. Static convolution proves effective in low-capacity models (around 4.5M parameters) but lags behind in larger-scale models. Additionally, we observe that the inter-window information transfer in local-attention STMs significantly influences performance.

Moreover, our aim extends beyond performance comparison to uncover distinctive characteristics in different STMs. 
The design of an STM, reflecting prior knowledge and constraints in the hypothesis space (inductive bias), is analyzed for its impact on learned model characteristics, including locality, shift, rotation, and scale invariance.
Observations reveal significant variations among STMs in capturing information during feature aggregation, termed as the effective receptive field (ERF)~\cite{luo2016understanding}. Further exploration highlights that a larger ERF does not guarantee improved performances, and the benefits may saturate as the model scales up.
In our invariance test, models exhibiting superior performance also demonstrate heightened robustness against variations, suggesting the learnability of invariance through data and diverse augmentations. Besides, static convolution, employing weight-sharing and emphasizing locality, shows more pronounced translation invariance. Notably, using a flexible sampling strategy for dynamic aggregation, dynamic convolution (DCNv3~\cite{2023intern}) outperforms other STMs in rotation and scaling invariance. 
Besides, halo attention and DCNv3 also show superior adversarial robustness.

\begin{table*}[tp]
	\centering
	\caption{{Comparison of various spatial token mixer designs.}}
 \vspace{-2mm}
% \vspace{-3mm}
	\small
    \resizebox{0.98\linewidth}{!}{%
    \begin{tabular}{cccccc}
    \toprule
    STM                         & Type                & Abbr.     & Sampling Point Set                     & Aggregation Weight                    & Method                \\ \midrule
    Halo Attention              & Local Attention     & Halo-Attn & Local  & Dynamic              & HaloNet~\cite{vaswani2021scaling}                \\
    Spatial Reduction Attention & Global Attention    & SR-Attn   & Global     & Dynamic              & PVT~\cite{wang2022pvt} \\
   Shifted Window (Swin) Attention    & Local Attention     & SW-Attn & Local           & Dynamic              & Swin Transformer~\cite{liu2021swin}       \\
    Depth-Wise Convolution      & Convolution         & DW-Conv   & Local         & Static          & ConvNeXt~\cite{liu2022convnet}, DWNet~\cite{han2021connection}    \\
    Deformable Convolution v3     & Dynamic Convolution & DCNv3    & Global   & Dynamic & InternImage~\cite{2023intern}             \\ \bottomrule
    \end{tabular} }
	\label{tab:STM-comparison}
    \vspace{-4mm}
\end{table*}

 %\sm{Our goal partially resembles a concurrent work MetaFormer~\cite{DBLP:conf/cvpr/YuLZSZWFY22}, which also distills the advantageous engineering techniques from transformers and finds that transformer's block design contributes greatly to the performance gain. They conclude the concept of the spatial token mixer and also elaborate a unified architecture that yields competitive performance with simple STMs such as pooling and identity mapping~\cite{yu2022metaformer}. Compared with Metaformer which focuses on the unified architecture, our work aims to produce a comprehensive comparison of existing popular STMs on an advanced, unified architecture to reveal their real performance gap and characteristics behind the performance. Besides, our exploration of the unified architecture is not limited to the block design principle. Ablations on stage-level design and the comparison with the ConvNeXt~\cite{liu2022convnet} block are conducted. Hence, this work can be viewed as complementary to Metaformer, where more STMs are compared and analyzed, and more ablation on the meta-architecture is made.}

Our contributions can be summarized as follows:

\begin{itemize}[]
	
	\vspace*{-0.25mm}
    \item We experimentally compare typical spatial token mixers, unveiling the characteristics of local attention, global attention, depth-wise convolution, and dynamic convolution. 

    \vspace*{-0.25mm}
    \item Distilling effective network engineering techniques from modern backbone designs, we create a unified architecture that enhances spatial token mixer performance compared to their original implementations.

    \vspace*{-0.25mm}
	 \item  Our experiments using the unified architecture on image classification and object detection tasks reveal significant performance improvements due to network engineering techniques. Despite this, performance gaps persist among different STMs, with the earlier halo attention achieving state-of-the-art results.

     \vspace*{-0.25mm}
     \item
      Analyzing the impact of inductive bias on STM characteristics, we challenge the notion that the effective receptive field alone is a metric for evaluating STM quality. Our findings underscore that STMs with strong performance exhibit high invariance, emphasizing the role of STM design in contributing to this invariance.

       \vspace*{-0.25mm}
     \item All models and codes used in this study are publicly available at \url{https://github.com/OpenGVLab/STM-Evaluation}.

\end{itemize}

%% file: Section2.tex
\section{Related Work}
\subsection{Convolution-Based Spatial Token Mixers} 

Leveraging image-specific priors, such as sparse interactions, parameter sharing, and translation equivalence~\cite{Goodfellow-et-al-2016}, convolution has earlier occupied vision backbones~\cite{krizhevsky2017imagenet,simonyan2014very,szegedy2015going}.
The evolution of vision backbones primarily centered on architectural aspects, including residual learning~\cite{he2016deep}, dense connections~\cite{huang2017densely}, grouping techniques~\cite{howard2017mobilenets,xie2017aggregated}, and channel attention~\cite{hu2018squeeze}. 
In contrast, spatial feature aggregation mechanisms received less attention.
Approaches like deformable convolution~\cite{dai2017deformable} and non-local methods~\cite{wang2018non} introduced flexible point sampling and long-range dependency, breaking the locality constraint. 
With the emergence of vision transformers~\cite{dosovitskiy2020image}, convolution has adapted to more advanced network engineering designs~\cite{liu2022convnet,ding2022scaling,liu2022more}.
Notably, Han \textit{et al.}~\cite{han2021connection} demonstrated that introducing dynamic weight computation into depth-wise convolution can achieve comparable or slightly superior performance compared to local vision transformers.

\subsection{Attention-Based Spatial Token Mixers} 

Since the introduction of the transformer into vision~\cite{dosovitskiy2020image}, its core operator, attention, has significantly challenged the dominance of convolution. 
Unlike standard convolution, attention boasts a global receptive field and dynamic spatial aggregation.
However, the dense attention calculation comes with a substantial computational overhead, scaling quadratically with the size of input maps. 
This becomes particularly impractical for applying vanilla attention to feature pyramids, especially on large feature maps. 
Consequently, attention operators in vision typically reduce the number of locations attended during attention calculation.
Approaches like PVT~\cite{wang2021pyramid,wang2022pvt} and Linformer~\cite{wang2020linformer} employ global attention on down-sampled feature maps, while Pale Transformer~\cite{wu2022pale} uses pale operation to sample features across the entire feature map.
Inspired from the locality prior of convolution, local attention is introduced~\cite{vaswani2021scaling,liu2021swin}, where attention is constrained within local windows. Since these local windows are non-overlapping, mechanisms for inter-window information transfer, such as ``haloing''~\cite{vaswani2021scaling} and shifted windows~\cite{liu2021swin}, have been developed.

\subsection{Optimizing Engineering and Backbone Evaluation}
Instead of introducing novel spatial token mixers, some research focuses on identifying advantageous network engineering techniques.
ResNet-RSB~\cite{wightman2021resnet} demonstrates that an advanced training recipe can significantly enhance performance compared to a na\"ive setting.
ConvNeXt~\cite{liu2022convnet} modernizes a standard ResNet by incorporating recent advancements in both training recipes and network architecture design.
MetaFormer~\cite{DBLP:conf/cvpr/YuLZSZWFY22} underscores the significant role of the abstracted architecture of the Transformer block in achieving competitive performance.
Recent work by Yu \textit{et al.}~\cite{yu2022metaformer} further validates the effectiveness of Transformer-style block design, achieving impressive performance using basic or common mixers.
Despite MetaFormer's emphasis on the importance of the transformer block, diverting attention from STMs, we observe performance gaps among different STMs.
Toolkit initiatives like Timm~\cite{rw2019timm} offer off-the-shelf or reproduced models but lack a comprehensive comparison of different STMs under a unified architecture.
As depicted in Fig.~\ref{fig:teaser}, an advanced STM design, such as halo-attention, coupled with a modern network architecture, outperforms other STMs with the same architecture by a significant margin.
Additionally, we find that network engineering techniques extend beyond block-level design. We conduct more ablation studies on stage-level design and compare them with ConvNeXt's block design.
It's important to note that our primary goal is not to construct an extremely optimized framework but to comprehensively compare and analyze different STMs under a unified setting from various perspectives.

%% file: Section3-evolution.tex
\section{Spatial Token Mixer}
\label{sec:stm}

\subsection{Definition of Spatial Token Mixer}
The Spatial Token Mixer (STM) aims to aggregate spatial features around each pixel to transfer contextual information. Given an input feature map $\boldsymbol{X} \in \mathbb{R}^{C\times H\times W}$, we define the token mixing function $\mathcal{M}$ for the mixing target point $p$ as:
\begin{equation}
	\label{eq:spatial-token-mixer-definition}
	\mathcal{M} \left( p \right) =W_o\sum_{k\in S}{w_{pk}\cdot \boldsymbol{X}}\left(k \right) \ ,
\end{equation}
where $S$ denotes the \textit{sampling point set} that records the pixels to attend; ${w_{pk}}$ denotes the aggregation weight with respect to the sample point $p$; and
$W_o$ is the output projection matrix.
In practice, STMs typically employ a multi-head strategy, projecting or splitting the input feature into different heads.
STMs vary in the generation of the sampling point set and the aggregation weight, as summarized in Table~\ref{tab:STM-comparison}.
In this study, we consider four prevalent types of STMs for comparison: depth-wise convolution, dynamic convolution, local attention, and global attention.

\subsection{Taxonomy of Spatial Token Mixer}

\subsubsection{Depth-Wise Convolution}
Depth-wise convolution aggregates features in an input-independent manner. The sampling point set consists of sliding windows centered at the mixing target point $p$, and the aggregation weight is represented by a learnable weight matrix—the convolution kernel. Despite its simplicity, recent convolution networks~\cite{liu2022convnet,ding2022scaling} have demonstrated that depth-wise convolution, when combined with advanced architecture and training recipes, is able to achieve comparable performance to top-performing vision transformers. 
In our comparison, we employ a $7\times 7$ depth-wise convolution, following the configuration of DWNet~\cite{han2021connection} and ConvNeXt~\cite{liu2022convnet}. 
Specifically, we incorporate input and output projection layers before and after the depth-wise convolution in our unified block design.

\subsubsection{Dynamic Convolution}
Dynamic convolution is input-dependent, where both the sampling point set and aggregation weights are generated conditioned on the inputs.
For instance, the widely-used deformable convolution~\cite{dai2017deformable,zhu2019deformable,2023intern} infers per-pixel sampling point sets and aggregation weights from input features for spatial aggregation. 
In this comparison, we adopt the deformable convolution v3~(DCNv3) from InternImage~\cite{2023intern} as an example of dynamic convolution.
We employ nine sampling points, consistent with the original implementation. The offsets and weights for these sampling points are determined through a $3\times 3$ depth-wise convolution followed by a linear projection. Weights are normalized using a \texttt{Softmax} function.

\subsubsection{Global Attention}
Global-attention-based STM mixes features across the entire spatial domain, with aggregation weights dynamically generated through the inner product between the target point and each sampling point.
While global attention offers a theoretically global receptive field without introducing image-specific inductive bias like locality, it comes with a significant computational overhead that scales quadratically with input resolutions. 
To integrate global attention STMs into pyramid-like vision backbones, optimization for efficiency is essential. 
A common practice involves constraining the cardinality of the sampling point set~\cite{wang2021pyramid,dong2022cswin,wu2022pale}. 
In our experiments, we adopt the spatial reduction attention from PVT~\cite{wang2021pyramid}. During attention computation, the \textit{key} and \textit{value} feature maps are downsampled.

\subsubsection{Local Attention}
An alternative approach to optimizing global attention involves introducing locality, i.e., restricting the sampling point set to local windows~\cite{vaswani2021scaling,liu2021swin}, rather than the entire spatial domain.
Given the intricacy of attention, local attention, in contrast to the densely overlapped sliding windows in convolution, employs non-overlapping local windows where pixels within each window share the same sampling point set. 
To facilitate information transfer among different windows, HaloNet~\cite{vaswani2021scaling} enlarges each window with an additional band resembling a ``halo.'' 
Swin Transformer~\cite{liu2021swin} introduces shifted windows to create overlapping sampling points, enabling information transfer between different blocks.
For a comprehensive exploration of the design principles of local attention STMs, we include both halo attention from HaloNet and shifted window attention from Swin Transformer in our comparison.

\begin{figure}[!t]
	\centering
	\includegraphics[width=0.88\linewidth]{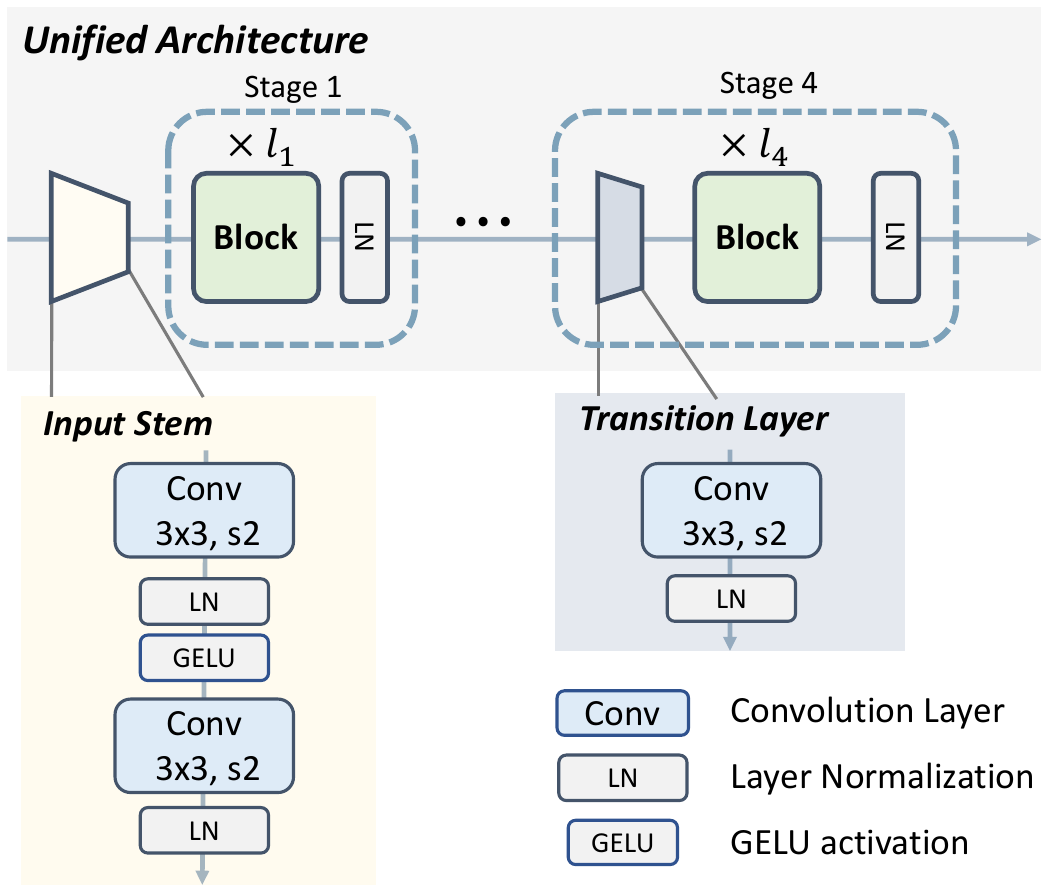}
	\caption{Unified architecture design. Overlapped downsampling is applied to the input stem and transition layer, while the standard transformer~\cite{vaswani2017attention} block design is adopted.}
	\label{fig:stage-design}
 \vspace{-3mm}
\end{figure}

%% file: Section4.tex
\section{Unified Architecture and Training Recipe}
This section presents our unified architecture and training strategies. We systematically analyze the stage-level and block-level architecture design through sequential ablations. Subsequently, we instantiate different STMs discussed in Sec.~\ref{sec:stm} into the unified architecture for comprehensive comparison.

\begin{table}[tp]
\setlength{\tabcolsep}{2pt}
\centering
\caption{The ablation study on block and stage design. }
\vspace{-2mm}
    \small
    \resizebox{0.8\linewidth}{!}{%
    \begin{tabular}{ccc|ccc}
    \toprule
    \multicolumn{3}{c|}{Stage Design} & \multicolumn{3}{c}{Block Design} \\
    STM       & Original & Unified & STM      & Original & Unified \\ \midrule
    SW-Attn      & 81.1     & 82.3       & Halo-Attn  & 80.5     & 83.0       \\
    DW-Conv  & 82.1     & 82.2       & DW-Conv & 82.2     & 82.2       \\ \bottomrule
    \end{tabular}
    \label{tab:block_stage} }
    \vspace{-3mm}
\end{table}

\subsection{Stage-Level Design}
First, we determine the stage-level design of the unified architecture, focusing on the input stem and transition layers. Recent backbones~\cite{liu2021swin,liu2022convnet,wang2021pyramid} often perform downsampling in the transition layer, with downsampling categorized into overlapped~\cite{DBLP:conf/cvpr/YuLZSZWFY22} (a $3\times3$ convolution with a stride of two) and non-overlapped~\cite{liu2021swin,liu2022convnet} (a $2\times2$ convolution with a stride of two) patch merging. Here, we compare the performance of these two downsampling methods.

As depicted in Table~\ref{tab:block_stage} (left part), the use of overlapped downsampling improves the classification accuracy of Swin-Transformer by $1.2\%$, while the improvement for ConvNeXt is marginal ($0.1\%$).
A plausible explanation is that overlapped downsampling facilitates information transfer among windows for local attention STMs, which complements the inherent weakness of non-overlapped window partition: the lack of inter-window information transfer. 
However, for the $7\times 7$ depth-wise convolution used in ConvNeXt, minor gains are observed, as the sliding windows of convolution are densely overlapped.
Hence, we adopt the $3\times3$ convolutions with a stride of two for the input stem and transition layers; refer to Fig.~\ref{fig:stage-design} for details.

\subsection{Block-Level Design}
\label{sec:block-level-design}
Block-level design determines the topology of operators, specifying their positions and connections.
Before the emergence of vision transformers, most vision models adhere to the block design of the standard ResNet~\cite{he2016deep}, featuring a single skip connection with standard convolution (as seen in the HaloNet block design in Fig.~\ref{fig:arc}).
The transformer block~\cite{vaswani2017attention,dosovitskiy2020image} introduces a different approach by segregating STM and channel feature transformation into two residual connections, each with a normalization layer (refer to the unified block design in Fig.~\ref{fig:arc}).
Metaformer~\cite{DBLP:conf/cvpr/YuLZSZWFY22} demonstrates that the transformer-style block design itself significantly contributes to the performance gain of vision transformers.

Hence, for the unified architecture, we adopt a standard transformer block as depicted in Fig.~\ref{fig:arc} and replace the attention with different types of STMs.
In the right part of Table~\ref{tab:block_stage}, the unified block design substantially enhances the performance of HaloNet ($+2.5\%$), which employs the ResNet-style block design in its original implementation.
However, ConvNeXt~\cite{liu2022convnet} achieves comparable performance with depth-wise convolution using an improved ResNet block design (denoted as ``original'' in Table~\ref{tab:block_stage}). 
This suggests that the transformer's block design may not necessarily be a universal rule. 
Despite this, for the sake of uniformity in architecture for comparison, we continue to use the unified block design for all compared STMs, including depth-wise convolution. 
We further compare the compatibility of the ConvNeXt block and the transformer block for different STMs in Sec.~\ref{sec:architecture-ablation}.
Here, we denote the overall network, where the depth-wise convolution used in the unified architecture, as ``U-DWConv''.

% Depth-wise convolution has been used in a ``modern'' CNN architecture  ConvNeXt~\cite{liu2022convnet}, which has a very different block design compared with the transformer block. In our work, considering the uniformity of architecture for comparison, we adopt this unified block design for all the compared STMs, including depth-wise convolution. 
% %
% Here, we denote ``Original'' as the block architecture used in ConvNeXt and find that ConvNeXt~\cite{liu2022convnet} with an improved Transformer's block design can attain comparable performance for the depth-wise convolution.
% %
% This suggests that the transformer's block design may not necessarily be a golden rule.
% %Some simple STMs may have more suitable block designs. 
% In our work, considering the uniformity of architecture for comparison, we adopt this unified block design for all the compared STMs, including depth-wise convolution. 
% %
% Here, we denote the overall network, where the depth-wise convolution used in the unified architecture, as the ``U-DWConv''.
% %
% We further compare the compatibility of the ConvNeXt block and the transformer block for different STMs in Sec.~\ref{sec:architecture-ablation}. 

\subsection{Model Configuration and Training Strategies}

\subsubsection{Model Configuration}
To facilitate a comprehensive comparison, we instantiate STMs on models with four different scales: micro~($\sim$4.5M), tiny~($\sim$30M), small~($\sim$50M), and base~($\sim$90M).
Implementations under our unified architecture are denoted with the prefix ``U-''.
Parameters and MACs of different models are detailed in Table~\ref{tab:imagenet1k-comparison}.
It is worth noting that micro models, with only around $4.5M$ parameters, are seldom discussed for standard backbones.
Given the differing complexities of various STMs, the blocks per stage and channel numbers also vary.
Further details can be found in the supplementary material.

\begin{table}[!t]
%\small
\caption{{Classification accuracy on ImageNet-1k.} ``U-'' denotes that the models are re-implemented in our unified architecture.}
\vspace{-2mm}
\resizebox{0.99\linewidth}{!}{%
    \begin{tabular}{c|l|r|r|c}
    \toprule
    Scale                  & \multicolumn{1}{c|}{Method} & \multicolumn{1}{c|}{\#Params} & \multicolumn{1}{c|}{FLOPs} & ImageNet-1k Acc \\ \midrule
    \multirow{5}{*}{Micro} & U-HaloNet                  & 4.4M                         & 0.65G                     & \textbf{75.8\%} \\
                           & U-PVT                      & 4.3M                         & 0.57G                     & 72.8\%          \\
                           & U-Swin Transformer                     & 4.4M                         & 0.71G                     & 74.4\%          \\
                           & U-DWConv                 & 4.4M                         & 0.65G                     & 75.1\%          \\
                           & U-InternImage              & 4.3M                         & 0.65G                     & 75.3\%          \\ \midrule
    \multirow{5}{*}{Tiny}  & U-HaloNet                  & 31.5M                        & 4.75G                     & 83.0\%          \\
                           & U-PVT                      & 30.8M                        & 4.56G                     & 82.1\%          \\
                           & U-Swin Transformer                     & 31.5M                        & 4.91G                     & 82.3\%          \\
                           & U-DWConv                 & 31.9M                        & 5.01G                     & 82.2\%          \\
                           & U-InternImage              & 29.9M                        & 4.83G                     & \textbf{83.3\%} \\ \midrule
    \multirow{5}{*}{Small} & U-HaloNet                  & 52.8M                        & 8.92G                     & 84.0\%          \\
                           & U-PVT                      & 50.5M                        & 7.33G                     & 83.2\%          \\
                           & U-Swin Transformer                     & 52.9M                        & 9.18G                     & 83.3\%          \\
                           & U-DWConv                 & 54.4M                        & 9.40G                     & 83.1\%          \\
                           & U-InternImage              & 50.1M                        & 8.24G                     & \textbf{84.1\%} \\ \midrule
    \multirow{5}{*}{Base}  & U-HaloNet                  & 93.3M                        & 15.84G                    & \textbf{84.6\%} \\
                           & U-PVT                      & 91.1M                        & 12.73G                    & 83.4\%          \\
                           & U-Swin Transformer                     & 93.4M                        & 16.18G                    & 83.7\%          \\
                           & U-DWConv                 & 95.8M                        & 16.64G                    & 83.7\%          \\
                           & U-InternImage              & 97.5M                        & 16.08G                    & 84.5\%          \\ \bottomrule
    \end{tabular}
    }
    \label{tab:imagenet1k-comparison} 
    \vspace{-3mm}
\end{table}

\subsubsection{Image Classification} 
We consolidate training techniques from recent top-performing networks~\cite{liu2021swin,liu2022convnet} into a unified strategy. 
Models undergo 300 training epochs on ImageNet-1k~\cite{deng2009imagenet} with the AdamW optimizer~\cite{loshchilov2017decoupled}. 
Training specifics include gradient clipping with a maximum norm of 5.0, weight decay of 0.05, layer scale~\cite{touvron2021going} starting at $10^{-6}$, label smoothing~\cite{szegedy2016rethinking} at 0.1, and stochastic depth~\cite{huang2016deep}. 
Standard data augmentation practices~\cite{liu2021swin,liu2022convnet} are employed, incorporating RandAugment~\cite{cubuk2020randaugment}, Mixup~\cite{zhang2017mixup}, Cutmix~\cite{yun2019cutmix}, Random Erasing~\cite{zhong2020random}, and color jitter.
Training settings are uniform across all models, except for batch size and drop path rate. 
Batch sizes are set to 1,024 for U-Swin Transformer, U-HaloNet, and U-PVT, and 4,096 for U-DWConv and U-InterImage, as larger batch sizes favor convolutional STMs. 
Drop path rates adhere to the original settings in their respective papers. 
The learning rate is determined as $(\text{batch size}/1024)\times10^{-3}$ with a linear warm-up lasting 20 epochs and cosine decay. 
Refer to Table~\ref{tab:imagenet1k-params} for detailed parameters.

\subsubsection{Object Detection} 
We employ object detection as the downstream task and integrate the models into Mask-RCNN~\cite{he2017mask} for evaluation. 
Models are trained and evaluated on COCO \texttt{train2017} and \texttt{val2017} with single-scale training and evaluation. 
Images are resized so that the shorter sides equal $800$ pixels while the longer sides do not exceed $1,333$ pixels. 
All models are trained using AdamW~\cite{loshchilov2017decoupled} for $12$ epochs with a batch size of $16$. 
The learning rate is set to $10^{-4}$ and drops by $10\times$ at the eighth and tenth epochs.
The weight decay is set to 0.05 and the warmup iterations are 1000. 
Drop path rates are set to $0.1$, $0.2$, $0.2$, and $0.3$ for micro, tiny, small, and base models, respectively.

\begin{table}[tp]
	\centering
	\caption{ImageNet-1k classification training settings.}
 \vspace{-2mm}
	\resizebox{0.85\linewidth}{!}{%
		\begin{tabular}{c|c|c}
			\toprule
			Training Configurations    &  Convolutions  &  Transformers  \\ 
			\midrule
			training epochs & 300 & 300 \\
			batch size & 4096 & 1024 \\
			warm-up epochs & 20 & 20 \\
			learning rate schedule & cosine decay & cosine decay \\
			base learning rate & 4e-3 & 1e-3 \\
			optimizer & AdamW & AdamW \\
			gradient clip & 5 & 5 \\
			weight decay & 0.05 & 0.05 \\
			layer scale & 1e-6 & 1e-6 \\
			label smoothing & 0.1 & 0.1 \\
			EMA decay & 0.9999 & 0.9999 \\
			input resolution & $224\times 224$ & $224\times 224$ \\
			mixup & 0.8 & 0.8 \\
			cutmix & 1.0 & 1.0 \\
			random erasing & 0.25 & 0.25  \\
			color jitter & 0.4 & 0.4 \\
			test crop ratio & 0.875 & 0.875 \\
			\bottomrule
		\end{tabular}
	}
	\label{tab:imagenet1k-params} 
	\vspace{-3mm}
\end{table}

%% file: Section5.tex
\section{Results and Discussion}

\subsection{Image Classification}

We present the comparison of different STMs on ImageNet-1k classification in Table~\ref{tab:imagenet1k-comparison}.
U-InternImage achieves the best performance on tiny and small models but slightly falls behind U-HaloNet on micro and base scales.
Halo-Attn demonstrates comparable performance with DCNv3 and outperforms SW-Attn in Swin Transformer significantly. This suggests that ``haloing'' is a practical window information transfer design for local-attention STMs when performance is a priority.
It highlights that the relatively early STM, such as halo attention, can yield a significant performance gain with advanced network engineering techniques.
Using the $7\times7$ DW-Conv can achieve comparable performance with SW-Attn and surpasses SW-Attn on micro models.
U-PVT with global attention (SR-Attn) as STM shows obvious disadvantages under micro models but quickly becomes comparable with U-Swin Transformer on other scales.

\subsection{Object Detection}

We use object detection as the down-stream task to evaluate different STMs. Table~\ref{tab:object-detection-comparison} reports the detection results, where (i) STMs with high accuracy (Halo-Attn and DCNv3) retain the advantages on object detection and surpass the other STMs by large margins; 
(ii) global attention (U-PVT) performs the worst on micro models but becomes on par with SW-Attn on other scales; 
(iii) depth-wise convolution shows strong performance among the micro models, but is not competitive at larger scales.
In our experiments, both local-attention- and convolution-based STMs with locality inductive bias perform better compared with the model using global spatial reduction attention. 

% Considering the results from both the up-stream and the down-stream tasks, both local-attention- and convolution-based models with locality inductive bias perform better compared with the model using global spatial reduction attention. 
%

\setlength{\tabcolsep}{2.5pt}
\begin{table}[!t]
    \small
    \centering
    \caption{{Object detection results on COCO.}}
    \vspace{-2mm}
    \resizebox{0.99\linewidth}{!}{%
\begin{tabular}{c|l|cccc|cccc}
\toprule
\multirow{2}{*}{Scale} & \multicolumn{1}{c|}{\multirow{2}{*}{Method}} & \multicolumn{4}{c|}{Box}                                      & \multicolumn{4}{c}{Mask}                                      \\
                       & \multicolumn{1}{c|}{}                        & $AP$          & $AP_s$        & $AP_m$        & $AP_l$        & $AP$          & $AP_s$        & $AP_m$        & $AP_l$        \\ \midrule
\multirow{5}{*}{Micro} & U-HaloNet                                    & \textbf{40.3} & \textbf{24.5} & \textbf{43.0} & \textbf{53.7} & \textbf{37.3} & \textbf{18.7} & \textbf{39.7} & \textbf{55.2} \\
                       & U-PVT                                        & 35.9          & 21.2          & 38.4          & 47.2          & 34.2          & 16.8          & 36.1          & 50.2          \\
                       & U-Swin Transformer                                       & 36.6          & 22.3          & 39.1          & 48.0          & 34.6          & 17.2          & 36.7          & 50.1          \\
                       & U-DWConv                                   & 39.2          & 23.0          & 42.6          & 51.0          & 36.4          & 17.8          & 39.2          & 52.7          \\
                       & U-InternImage                                & 39.5          & 22.8          & 42.9          & 52.7          & 36.6          & 17.4          & 39.4          & 53.6          \\ \midrule
\multirow{5}{*}{Tiny}  & U-HaloNet                                    & 46.9          & 30.0          & 50.2          & \textbf{61.4} & 42.4          & 22.6          & 45.3          & \textbf{60.9} \\
                       & U-PVT                                        & 44.2          & 27.5          & 47.4          & 59.1          & 40.6          & 21.1          & 43.4          & 59.5          \\
                       & U-Swin Transformer                                       & 44.3          & 28.0          & 47.5          & 57.8          & 40.5          & 21.6          & 43.0          & 58.2          \\
                       & U-DWConv                                   & 44.3          & 27.8          & 48.0          & 57.5          & 40.5          & 20.7          & 43.9          & 58.6          \\
                       & U-InternImage                                & \textbf{47.2} & \textbf{30.4} & \textbf{51.3} & 61.3          & \textbf{42.5} & \textbf{23.3} & \textbf{45.9} & \textbf{60.9} \\ \midrule
\multirow{5}{*}{Small} & U-HaloNet                                    & \textbf{48.2} & \textbf{31.2} & \textbf{52.5} & \textbf{63.2} & \textbf{43.3} & \textbf{23.4} & \textbf{47.0} & \textbf{62.3} \\
                       & U-PVT                                        & 46.1          & 28.8          & 49.3          & 61.2          & 41.9          & 21.9          & 44.6          & 61.2          \\
                       & U-Swin Transformer                                       & 46.4          & 28.2          & 50.0          & 61.5          & 42.1          & 21.8          & 45.2          & 60.7          \\
                       & U-DWConv                                   & 45.6          & 28.5          & 49.8          & 59.8          & 41.2          & 21.9          & 44.7          & 59.4          \\
                       & U-InternImage                                & 47.8          & 30.3          & 51.8          & 62.4          & 43.0          & 23.1          & 46.3          & 61.7          \\ \midrule
\multirow{5}{*}{Base}  & U-HaloNet                                    & \textbf{49.0} & 31.8          & \textbf{53.0} & \textbf{63.7} & \textbf{43.8} & 24.4          & \textbf{47.0} & \textbf{62.8} \\
                       & U-PVT                                        & 46.4          & 29.0          & 49.8          & 62.2          & 42.3          & 22.3          & 45.1          & 61.8          \\
                       & U-Swin Transformer                                       & 47.0          & 30.8          & 50.6          & 62.2          & 42.2          & 23.5          & 45.1          & 61.7          \\
                       & U-DWConv                                   & 46.7          & 30.2          & 50.3          & 61.4          & 42.2          & 22.8          & 45.1          & 60.7          \\
                       & U-InternImage                                & 48.7          & \textbf{33.1} & 52.5          & 63.4          & \textbf{43.8} & \textbf{25.0} & 46.9          & 62.3          \\ \bottomrule
\end{tabular}
    }
    \label{tab:object-detection-comparison} 
\end{table}

%However, the weight generation seems to be insensitive to model scales, local attention and deformable convolution perform competitively on micro models. 

\begin{figure*}[tp]
	\centering
	\includegraphics[width=0.95\linewidth]{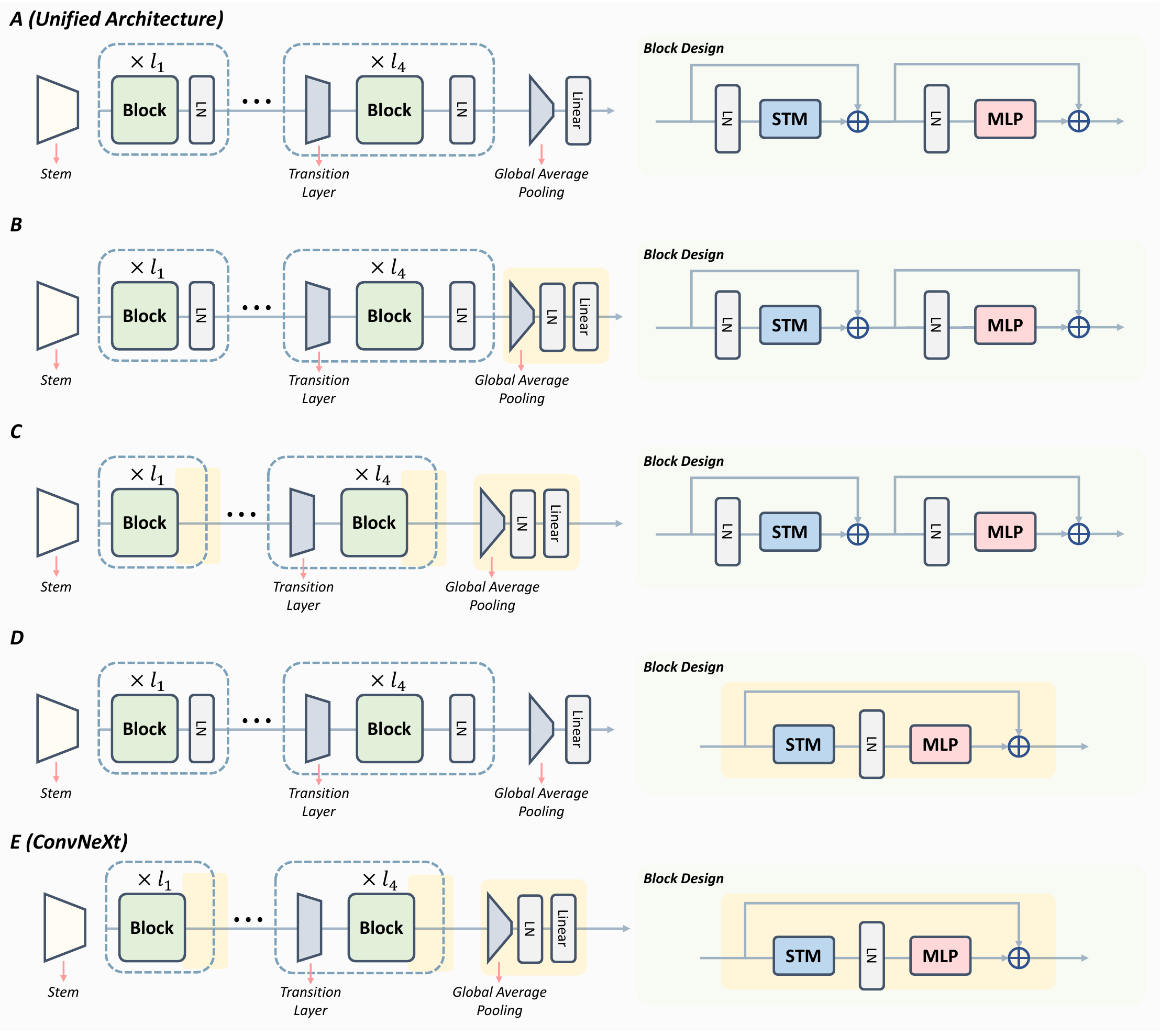}
	\caption{The illustration showcases five architectures, transitioning from our unified architecture (\textcolor[HTML]{FD6864}{A}) to the ConvNeXt architecture (\textcolor[HTML]{009901}{E}). The yellow shadings in architectures B-E emphasize their differences compared with architecture A.}
	\label{fig:supp-fig-1652}
	\vspace{-3mm}
\end{figure*}

\subsection{Ablation Study of the Unified Architecture}
\label{sec:architecture-ablation}

In this section, we conduct an ablation study on the unified architecture design for STMs to investigate how stage-level and block-level architecture designs affect the performance of different STMs. As discussed in Sec.~\ref{sec:block-level-design}, the $7\times7$ depth-wise convolution (DW-Conv) with the unified block design or ConvNeXt\cite{liu2022convnet} block design yields similar performance. This raises the question of whether ConvNeXt's block design can be an alternative to the transformer block design. 
Hence, we perform an ablation study for comparison, gradually transforming our unified architecture (\textcolor[HTML]{FD6864}{A}) into the ConvNeXt architecture (\textcolor[HTML]{009901}{E}). 

As shown in Fig.~\ref{fig:supp-fig-1652}, for the stage-level design, \textbf{B} shifts the layer normalization (LN) after the global average pooling in the decoder head based on A, and \textbf{C} removes LN after each stage based on B. For the block-level design, \textbf{D} adopts the block design of ConvNeXt (Fig.~\ref{fig:arc} (e)) based on A. 
Note that we keep our stem and transition layer (Fig.~\ref{fig:stage-design}) in all models.

The results in Table~\ref{tab:ablation-convnext} reveal that: (i) for most STMs, adding LN after global average pooling (B) and removing stage normalization (C) consistently improve performance; (ii) except for DW-Conv, applying other STMs to the ConvNeXt-style block degrades performance (A and D). In contrast, DW-Conv shows a performance gain with the ConvNeXt-style block design. However, for a fair comparison, we retain the transformer-style block design. 
We retain stage normalization because its importance for maintaining stability during training, especially in larger models, is demonstrated in studies by \cite{liu2022swin, 2023intern}.

\begin{table}[!]
	\centering
	\caption{{Ablation study of the unified network architecture.} Experiments are conducted on small-level models. We gradually modify our unified architecture (denoted by {\color[HTML]{FD6864} A}) into the architecture of ConvNeXt (denoted by {\color[HTML]{009901} E}). ``Original'' results are the reported results in their papers.}
    \vspace{-2mm}
	\small
	\resizebox{0.67\linewidth}{!}{%
		\begin{tabular}{c|cccccc}
			\toprule
			STM       & {\color[HTML]{FD6864} A} & B             & C             & D    & {\color[HTML]{009901} E} & Original \\ \midrule
			Halo-Attn & 84.0                     & 84.3          & \textbf{84.4} & 83.8 & 83.8                     & $--$     \\
			SR-Attn   & 83.2                     & \textbf{83.4} & \textbf{83.4} & 78.6 & 79.4                     & 81.7     \\
			SW-Attn   & 83.3                     & 83.5          & \textbf{83.7} & 82.8 & 82.9                     & 83.0     \\
			DW-Conv   & 83.1                     & 83.1          & 83.3 & 83.2 & \textbf{83.6}                    & 83.1     \\
			%DW-Conv   & 83.1                     & 83.1          & \textbf{83.3} & 82.6 & 82.4                     & 83.1     \\
			%DW-Conv*  & 82.4                     & 82.6          & 83.5          & 83.2 & \textbf{83.6}            & 83.1     \\
			DCNv3     & \textbf{84.1}            & \textbf{84.1} & 84.0          & 83.8 & 83.6                     & $--$     \\ \bottomrule
	\end{tabular} }
	\label{tab:ablation-convnext}
\end{table}

\begin{table}[tp]
	\centering
	\caption{Ablation study on local attentions in small models.}
  \vspace{-2mm}
	\resizebox{0.68\linewidth}{!}{%
		\small
		\begin{tabular}{ccc}
			\toprule
			Methods  & MACs & ImageNet-1k Acc  \\ \midrule
			U-Swin Transformer        & 9.16 & 83.3 \\
			U-HaloNet         & 9.75 & 84.0 \\
			U-HaloNet-Switch & 9.45 & 83.9 \\
			U-HaloNet-1px    & 9.32 & 83.8 \\
			U-HaloNet-Shift  & 9.75 & 82.7 \\ \bottomrule
	\end{tabular}  }
	\label{tab:ablation-local-attention}
\end{table}

\begin{table}[tp]
\caption{Inference speed evaluated on an NVIDIA GeForce RTX 3090 GPU (24GB). The throughput is reported.}
\vspace{-2mm}
\label{tab:inference_speed}
\centering
\resizebox{0.83\linewidth}{!}{%
\begin{tabular}{c|c|c|c|c}
\hline
Scale & Method & 256x256 & 512x512 & 1024x1024\\ \hline
\multirow{5}{*}{Micro} 
& U-HaloNet       & 3627 & 979  & 254 \\
& U-PVT        & 3755 & 841  & 149 \\
& U-Swin Transformer       & 5307 & 1415 & 366 \\
& U-DWConv   & 4299 & 1103 & 287 \\
& U-InternImage & 3018 & 765  & 197 \\ \hline
\multirow{5}{*}{Tiny}  
& U-HaloNet       & 1133 & 330  & 86  \\
& U-PVT        & 1146 & 269  & 54  \\
& U-Swin Transformer       & 1587 & 460  & 125 \\
& U-DWConv   & 1582 & 397  & 102 \\
& U-InternImage & 818  & 208  & 53  \\ \hline
\multirow{5}{*}{Small} 
& U-HaloNet       & 663  & 200  & 54  \\
& U-PVT        & 827  & 193  & 38  \\
& U-Swin Transformer       & 913  & 275  & 76  \\
& U-DWConv   & 995  & 249  & 65  \\
& U-InternImage & 588  & 148  & 38  \\ \hline
\multirow{5}{*}{Base}  
& U-HaloNet       & 448  & 140  & 38  \\
& U-PVT        & 552  & 127  & 24  \\
& U-Swin Transformer       & 605  & 189  & 53  \\
& U-DWConv   & 740  & 185  & 47  \\
& U-InternImage & 419  & 106  & 27  \\ \hline
\end{tabular} }
\end{table}

\subsection{Ablation Study of Local Attention}
Local attention has gained popularity as a STM for vision models due to its notable performance and efficiency. 
However, our study reveals a significant performance gap among different implementations of local attention, such as Halo-Attn~\cite{vaswani2021scaling} and Swin-Attn~\cite{liu2021swin}. This observation motivates us to conduct an in-depth analysis to identify the core factors affecting performance.

The main differences between Halo-Attn and Swin-Attn can be attributed to three factors: 
(i) Halo-Attn conducts inter-window feature transfer once every block, more frequently than Swin-Attn;
(ii) Halo-Attn employs a larger window (sampling point set) with the extra ``halo'';
(iii) Halo-Attn shows a relatively isotropic feature aggregation pattern, while Swin-Attn tends to aggregate information from a particular direction (see Fig.~\ref{fig-erf-visualization}).
To investigate the impact of these factors, we introduce three new variants for Halo-Attn, as shown in Table~\ref{tab:ablation-local-attention}. 
First, U-HaloNet-Switch is added, conducting inter-window information transfer every two blocks.
Next, we introduce U-HaloNet-1px, with a halo size set to one.
Finally, we shift the query window of Halo-Attn to the top-left corner while keeping the total window size unchanged to examine the effect of non-central aggregation.

Our experimental results in Table~\ref{tab:ablation-local-attention} demonstrate that the impact of window size and inter-window transfer frequency is not significant: halving the inter-window transfer frequency reduces MACs by $0.30$G while still preserving performance, and reducing the halo size to one only results in a $0.2\%$ accuracy drop.
However, U-HaloNet-Shift shows a significant performance drop of $1.3\%$ on accuracy, suggesting that the geometry of inter-window interaction is the crucial factor for local attention operators.

\subsection{Inference Speed}

We evaluate the inference speed of compared models on an NVIDIA GeForce RTX 3090 GPU (24GB). Table~\ref{tab:inference_speed} presents the throughput (T) for different scales and resolutions. The throughput is measured for images of sizes 256x256, 512x512, and 1024x1024, with batch sizes of 128, 32, and 12, respectively.
The results reveal that U-Swin Transformer consistently achieves the highest throughput across all scales and resolutions, particularly excelling at the Micro and Tiny scales. U-HaloNet and U-PVT show moderate throughput, maintaining consistent performance across different resolutions. U-InternImage generally exhibits lower throughput but remains competitive and a faster version is recently publicly available~\cite{xiong2024efficient}. These results highlight the importance of selecting methods based on specific application requirements, balancing speed with other metrics.

%% file: Section6.tex
\section{Inductive Bias, Invariance, and Adversarial Robustness Analysis}
In this section, we explore the impact of the inductive bias of various STMs on learned model characteristics. We conduct analyses on effective receptive fields, translation invariance, rotation invariance, scaling invariance, and adversarial robustness.

\subsection{Effective Receptive Field}
\label{sec:erf-analysis}
The effective receptive field (ERF)\cite{luo2016understanding} of a unit in vision backbones refers to the region where the pixels influence the output of this unit. 
While the inductive bias introduced by different STMs sets the theoretical upper limit for the receptive field, the actual ERFs can vary. 
We employ the ERF toolbox~\cite{luo2016understanding} to assess the ERF of the center pixel in feature maps across different stages. 
Models trained on ImageNet-1k are used to compute the ERF maps. 
The ERF toolbox provides a gradient norm map, with each value indicating the impact of specific locations on the center point. By placing a gradually expanding window at the center, we determine the square length-to-input size ratio when the sum of the gradient norm within the window reaches $50\%$ of the image size. 
This ratio, denoted as ERF@50, serves as the quantitative metric following~\cite{ding2022scaling}.

\begin{figure}[!t]
	\centering
	\includegraphics[width=0.99\linewidth]{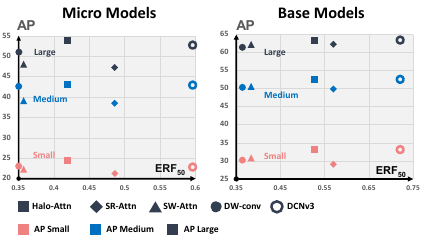}
	\caption{The correlation between detection accuracy and effective receptive field. We depict the box Average Precision (box AP) for small, medium, and large objects on the COCO validation set in relation to the ERF@50 of micro and base models.}
	\label{fig:erf-detection}
 \vspace{-3mm}
\end{figure}

We explore the relationship between Effective Receptive Field (ERF) and downstream performance in object detection. 
In Fig.~\ref{fig:erf-detection}, we present the box accuracy for small, medium, and large objects. Interestingly, we observe no strong correlation between ERF and detection accuracy. 
For instance, U-HaloNet exhibits comparable detection accuracy with smaller ERFs compared to U-InternImage. 
In the case of base models, accuracy comparisons across different scales remain consistent, and performance appears uncorrelated with ERF sizes.
Therefore, we posit that ERF should not serve as a definitive metric for evaluating STM quality. A smaller ERF does not necessarily imply inferiority compared to STMs with larger ERFs.

\begin{figure}[!t]
	\centering
	\includegraphics[width=0.99\linewidth]{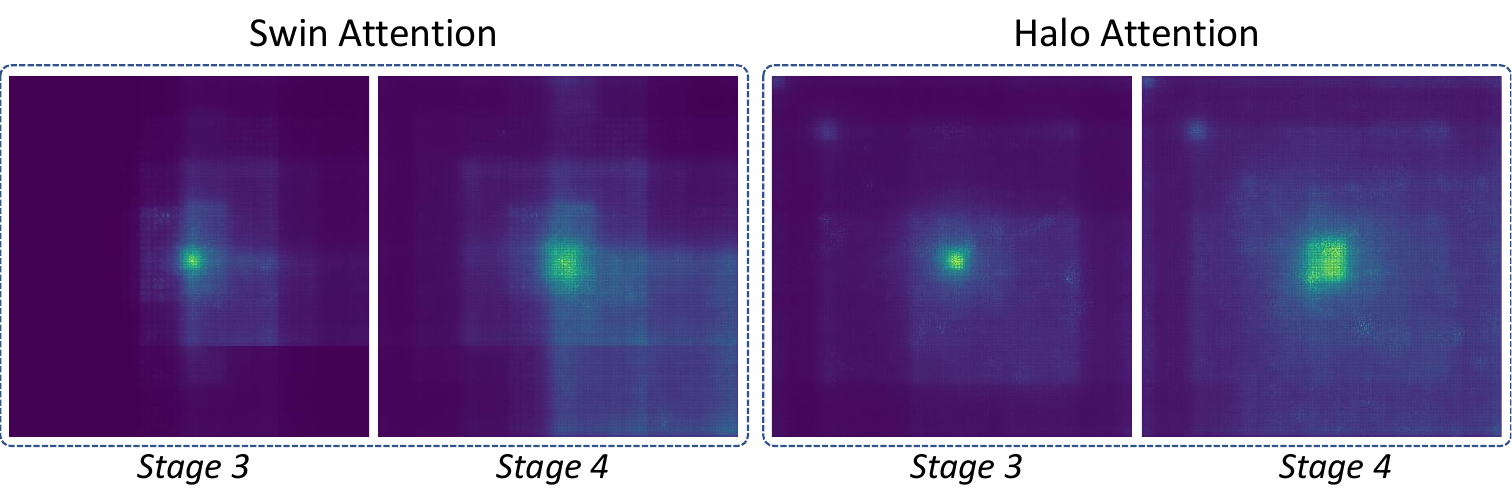}
	\caption{The design of the spatial token mixer influences the shape of the ERF. We visualize the ERF maps at the last two stages of U-Swin Transformer and U-HaloNet. The asymmetric ERF of Swin attention is evident, showcasing the presence of shifted windows.}
	\label{fig-erf-visualization}
 \vspace{-3mm}
\end{figure}

Furthermore, our observations indicate that the design of STMs influences the shape of the effective receptive field. 
Fig.~\ref{fig-erf-visualization} showcases visualized ERFs of SW-Attn and Halo-Attn. 
Notably, the shifted windows in the ERF of Swin-Attn are evident, with more features aggregated from the bottom-right corner, aligning with the direction of the shifted window. 
Conversely, halo attention exhibits a more symmetrical ERF due to the ``halo'' window information transfer. 
Additional results and visualizations of ERFs can be found in the supplementary material.

\subsection{Invariance Analysis}
\label{sec:invariance-analysis}
In this subsection, we assess the robustness of different STMs under various geometric transformations. We specifically consider translation, rotation, and scaling for our analysis. Invariance can be inherent in the STM design, learned from training samples, or acquired through data augmentation.
Convolution-based STMs inherently possess translation equivalence due to locality and weight sharing~\cite{Goodfellow-et-al-2016}, while local-attention STMs maintain window-level translation equivalence. The translation equivalence closely relates to translation invariance, given that the feature map undergoes average pooling before entering the classifier in our case.
Neither convolution-based nor attention-based STMs inherently incorporate inductive biases related to rotation or scaling.

\subsubsection{Translation Invariance}
Translation invariance refers to the model's ability to maintain consistent outputs when input images undergo translation. 
We assess translation invariance in the classification task by introducing jitter to the images, ranging from $0$ to $64$ pixels. 
Following~\cite{zhang2019making}, we measure invariance by the probability that the model predicts the same label for the same input images when translated.

\begin{figure}[!t]
	\centering
	\includegraphics[width=0.99\linewidth]{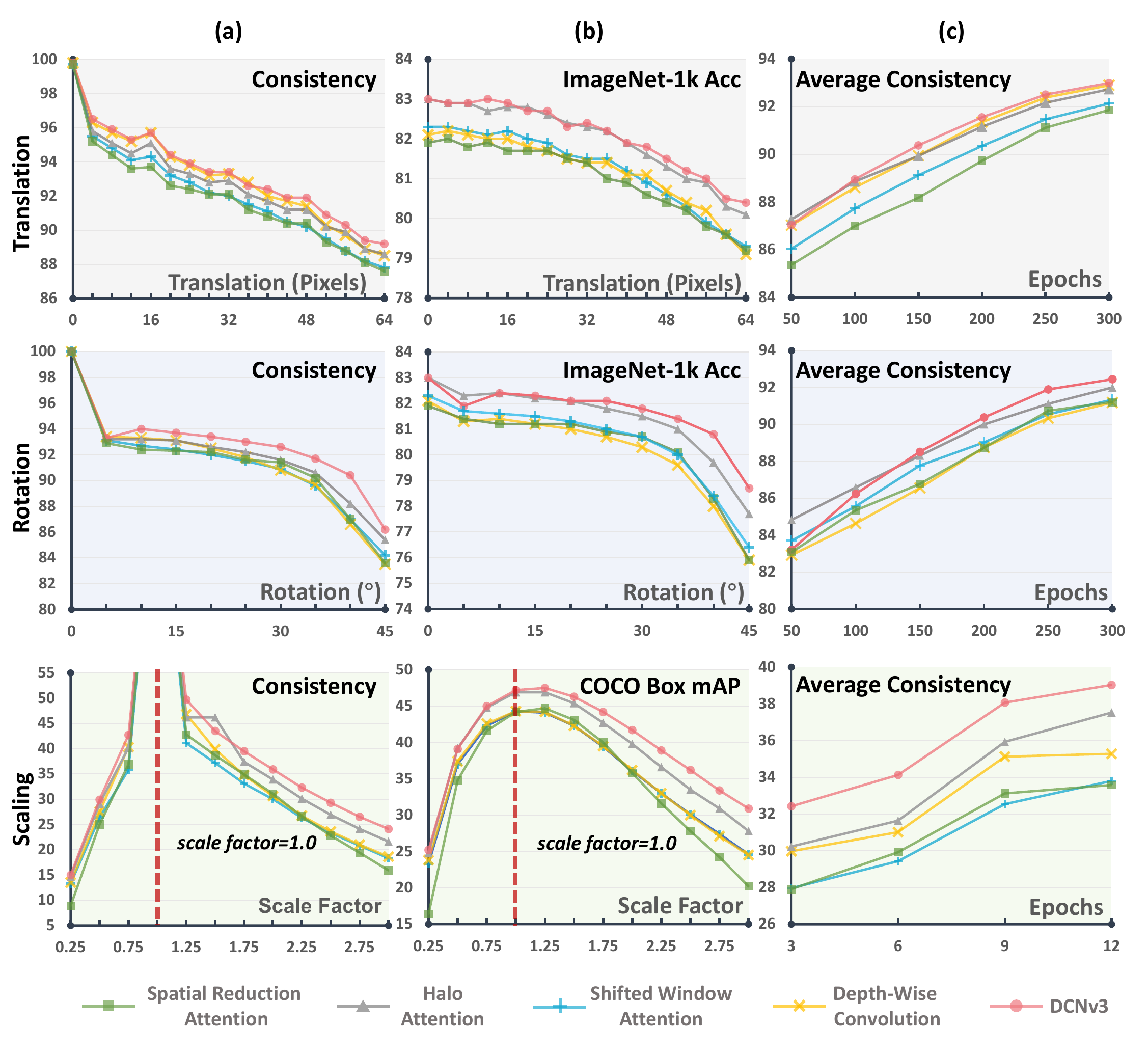}
	\caption{Translation, rotation, and scaling invariance of STMs. (a) Prediction consistency. (b) Classification (detection) accuracy under input translation, rotation, and scaling. (c) Average prediction consistency at different training epochs.}
	\label{fig:invariance-comparison-curve}
 \vspace{-3mm}
\end{figure}

In the first row of Fig.\ref{fig:invariance-comparison-curve}, we observe the translation invariance of different STMs, noting that:
(i) Convolution-based STMs (DCNv3 and depth-wise convolution) exhibit better translation invariance (see the first row in Fig.\ref{fig:invariance-comparison-curve} (a)).
(ii) U-PVT, using global attention, performs the poorest in terms of translation invariance, likely due to the absence of translation equivalence.
(iii) DCNv3 and halo attention consistently achieve the best performance even with translated images (see the first row in Fig.\ref{fig:invariance-comparison-curve} (b)).
(iv) Halo attention, depth-wise convolution, and DCNv3 demonstrate superior translation invariance initially, with DCNv3 improving its invariance during training (see the first row in Fig.\ref{fig:invariance-comparison-curve} (c)).

\subsubsection{Rotation Invariance} 
We evaluate rotation invariance on the classification task by rotating the images from $0\degree$ to $45\degree$ with a step of $5\degree$. Similar to translation invariance, we measure prediction consistency under different rotation angles to assess the rotation invariance of different STMs.

From the second row in Fig.~\ref{fig:invariance-comparison-curve}, we observe that:
(i) DCNv3, with its input-dependent sampling strategy, exhibits strong robustness against rotation.
(ii) Other STMs show comparable performance, with halo attention performing slightly better at larger rotation angles ($>30\degree$).
(iii) Rotation invariance is comparable among all STMs at the beginning of the training process, except for Halo-Attn, which demonstrates superior rotation invariance.
However, DCNv3 quickly learns strong rotation invariance from the training data.

\subsubsection{Scaling Invariance}

Scaling invariance is assessed in object detection by scaling input images with factors from $0.25$ to $3.0$ in steps of $0.25$. Box consistency serves as the invariance metric, where predicted boxes on scaled images are reverted to the original resolutions. Subsequently, the boxes predicted at the original resolutions are employed as ground truth boxes to compute the box mAP.

From the third row in Fig.~\ref{fig:invariance-comparison-curve}, we observe that:
(i) All STMs are sensitive to downscaling and exhibit comparable invariance with the input in small sizes.
(ii) DCNv3 performs better when upscaling the image, with both the box consistency and box mAPs outperforming the others.
(iii) Halo-Attn shows comparable invariance with depth-wise convolution in the early training stage but learns better invariance in the last three epochs.

\subsubsection{Summary}
Considering the results from translation, rotation, and scale invariance, we can conclude that STMs with strong performance exhibit relatively high invariance. 
However, the design of STMs also contributes to invariance; for example, depth-wise convolution attains comparable translation invariance with Halo-Attn.
Moreover, DCNv3, which learns to generate the sampling point set conditioned on inputs adaptively, achieves the best invariance among all the STMs. 
This suggests the generalization and robustness of DCNv3 compared to other STMs with fixed sampling sets.

Based on the study of inductive bias and invariance, we find the following guidelines for designing future STMs. First, incorporate inductive biases that support robustness to various transformations, such as translation, rotation, and scaling, as these have proven effective in enhancing model stability. Second, leverage both local and global feature aggregation to capture detailed spatial information and long-range dependencies, respectively, as this combination has shown to improve performance in various vision tasks. Finally, ensure efficient information transfer mechanisms within local windows to maximize the benefits of local receptive fields and improve overall model robustness and performance. 

\subsection{Adversarial Robustness Analysis}

 Adversarial robustness refers to the resilience of a model against adversarial attacks, which are inputs specifically designed to deceive the model into making incorrect predictions. We evaluate the adversarial robustness of the compared models on ImageNet. Following \cite{goodfellow2014explaining}, we use the fast gradient sign method (FGSM) to generate adversarial examples. Each model is trained on the clean dataset. FGSM is implemented to generate adversarial examples by computing the gradient of the loss function with respect to the input and creating perturbed inputs using various perturbation strengths ($\epsilon$=8/255, $\epsilon$=2/255, $\epsilon$=0.5/255). We evaluate each model's clean accuracy on the unperturbed test set and its adversarial accuracy on the generated adversarial examples for each $\epsilon$.

\begin{figure}[!t]
	\centering
	\includegraphics[width=0.99\linewidth]{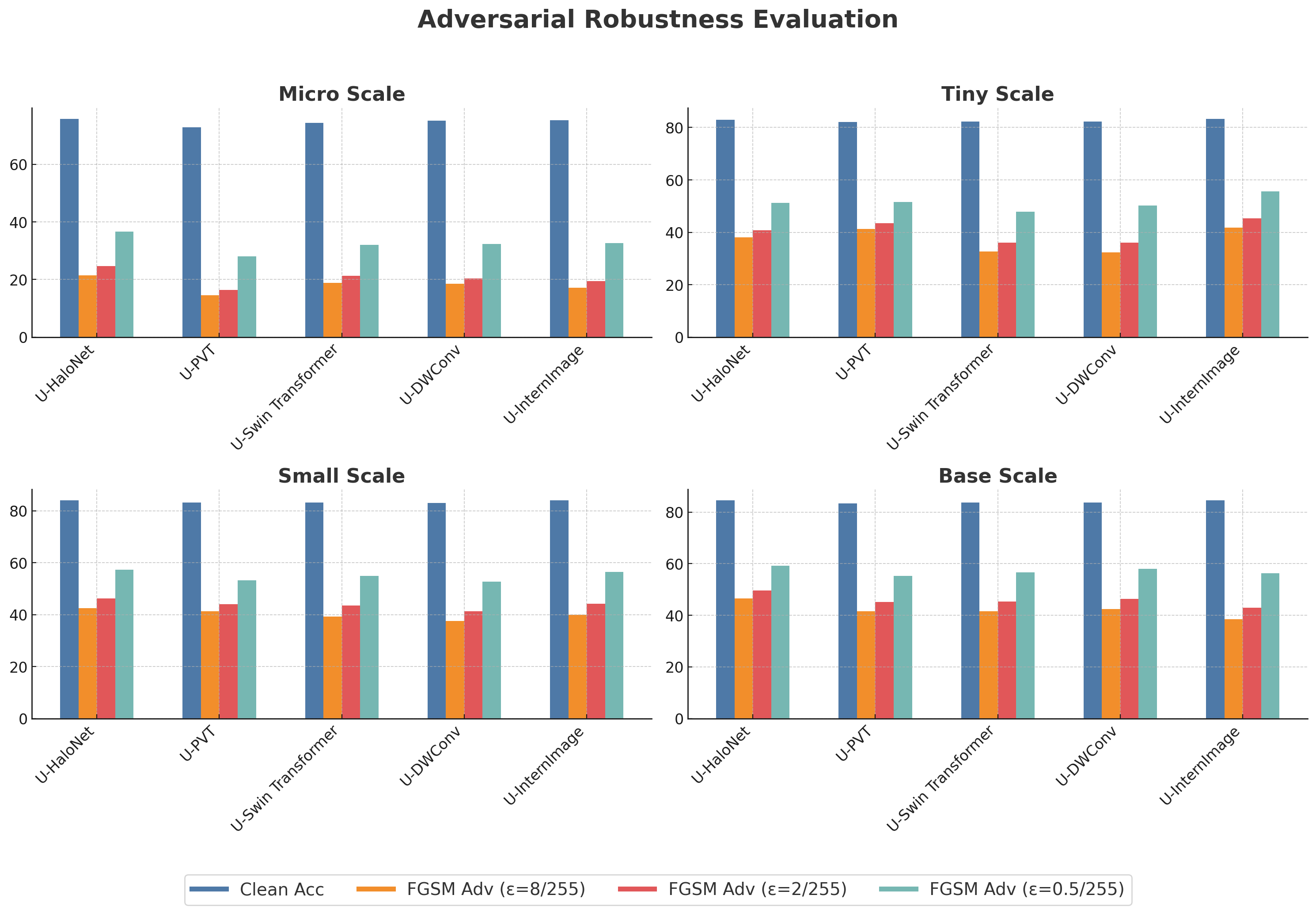}
	\caption{Accuracy on ImageNet-1k with adversarial attacks.}
	\label{fig:adversarial_robustness}
 \vspace{-3mm}
\end{figure}

 As shown in the Fig.~\ref{fig:adversarial_robustness}, our results reveal that (i) U-HaloNet using local attention consistently demonstrates moderate to high robustness across different scales, particularly excelling at higher perturbation strengths; (ii) U-PVT using global attention performs the worst in terms of adversarial robustness across all scales; and (iii) certain STMs, such as U-HaloNet and U-InternImage, demonstrate superior adversarial robustness, which is critical for AI safety and alignment.

%% file: Section7.tex
\section{Discussion}

\noindent \textbf{Generative tasks.}
The core principles and mechanisms that make STMs effective in discriminative models can indeed enhance generative models as well, given that both rely on efficient feature transformation and aggregation. STMs excel at aggregating spatial features, which is crucial for high-quality image synthesis in generative models. Additionally, integrating STMs with inherent biases like translation and scale invariance can lead to more stable and consistent outputs in generative tasks. Formulating a unified architecture to compare STMs on generative models across various tasks is an important area for future research.

\vspace{1.5mm}
\noindent \textbf{Mamba.} 
The Mamba architecture is designed for efficient sequence data processing. The first Mamba~\cite{gu2023mamba} model uses Selective State Space Models (SSMs) with input-dependent dynamics to enhance content-based reasoning in various modalities, achieving linear scaling with sequence length and higher throughput during inference. 
Mamba v2~\cite{dao2024transformers} introduces the Structured State Space Duality (SSD) mechanism, which bridges SSMs and attention mechanisms through structured semiseparable matrices. This results in the Mamba v2 architecture, featuring a more advanced block structure and multihead patterns. Consequently, Mamba v2 achieves faster performance while maintaining competitiveness with Transformers in language modeling.
Vision Mamba (ViM)~\cite{zhu2024vision} extends to visual representation learning by using bidirectional SSMs for rich visual context modeling and carefully designed block structures. ViM excels with subquadratic-time computation and linear memory complexity, making it highly efficient for high-resolution images.
Our study’s focus on spatial feature aggregation and the effectiveness of different STMs offers valuable insights when considering Mamba’s architecture. The performance of STMs is heavily influenced by their spatial feature aggregation strategies. Mamba’s SSMs, which use selective propagation and forgetting mechanisms, align well with our findings that effective spatial token mixing is crucial for high performance. Additionally, techniques such as bidirectional state space models and block-level normalization layers in Vision Mamba further enhance Mamba’s performance in the vision domain. We believe that our discussion outlines potential areas where Mamba could benefit from similar detailed examination.

%The recent Mamba~\cite{gu2023mamba} architecture utilizes selective state space models to efficiently manage information propagation. This enhances the model’s ability to handle long-range dependencies while maintaining computational efficiency. These enhancements have the potential to significantly improve the performance and stability of both discriminative and generative models. 

\vspace{1.5mm}
\noindent \textbf{Spiking neural networks.} SNNs~\cite{ghosh2009spiking} are a type of artificial neural network that more closely mimic the behavior of biological neurons compared to traditional neural networks. In SNNs, neurons communicate through discrete spikes or action potentials, with information encoded in the timing and frequency of these spikes. Neurons integrate incoming spikes, and when their membrane potential reaches a threshold, they emit a spike and reset. This process offers potential advantages in energy efficiency and temporal information processing. However, the spiking operation does not work well for vision tasks compared to convolution and attention operators due to its sparse and binary nature, which limits information density and complicates gradient calculation. SNNs struggle with efficient feature extraction and spatial relationship handling, which are critical for vision tasks. Additionally, the training algorithms and hardware support for SNNs are less mature, making them less effective and scalable compared to the well-established CNNs and attention-based models.

\vspace{1.5mm}
\noindent \textbf{Capsule networks.} Capsule networks~\cite{sabour2017dynamic} are designed to capture spatial hierarchies and pose information more effectively than traditional convolutional networks by grouping neurons into capsules that output vectors representing object properties. Dynamic routing in capsule networks is an iterative process where lower-level capsules send their outputs to higher-level capsules based on the agreement between their predictions and the actual outputs of the higher-level capsules.
However, dynamic routing introduces high computational complexity, poor scalability, and training instability due to the iterative updates, making it less effective for vision tasks compared to convolution and attention mechanisms. Convolutional networks, with their hierarchical pattern capture and efficient implementations, and attention mechanisms, with their ability to handle long-range dependencies and global context, are more data-efficient and benefit from extensive optimization and mature implementations. These advantages make convolution and attention operators more practical and effective for vision tasks than dynamic routing in capsule networks.

\section{Conclusion}

In this study, we compare popular Spatial Token Mixers (STMs) within vision backbones by establishing a unified setting to isolate the impact of network engineering techniques. Through ablation studies, we distill successful network engineering strategies and formulate a unified architecture, integrating five representative STMs for thorough comparison. Our experiments, encompassing both upstream and downstream tasks, scrutinize the influence of inductive bias on modern image deep networks.
Our key findings are as follows: (i) network engineering design is crucial for achieving performance gains, emphasizing the importance of thoughtful architecture choices; (ii) certain STMs, even those considered relatively early, can achieve state-of-the-art performance when coupled with modern design principles.
This work aims to provide guidelines for developing new vision backbones, particularly those integrating advanced spatial feature aggregation mechanisms. The results and analyses of the compared STMs offer valuable insights for practitioners, aiding in the judicious selection of suitable backbones for specific applications.

Although our study offers comprehensive information, we acknowledge certain limitations and outline directions for future research. First, we will evaluate STMs within larger models and diverse datasets to improve generalizability and robustness. Second, we plan to broaden the scope of the evaluation to include a wider range of downstream tasks and develop a unified architecture to compare STMs on generative models. Third, our current unified framework, while effective, may not fully optimize the unique strengths of each STM, particularly for future advanced models; future work will explore more tailored integrations for various STMs and tasks.